\documentclass[letterpaper, 10 pt, conference]{ieeeconf}  

\IEEEoverridecommandlockouts                              
\usepackage[utf8]{inputenc}
\usepackage[T1]{fontenc}
\usepackage{graphicx}
\usepackage{float} 
\usepackage{ragged2e}
\usepackage{booktabs}
\usepackage{bbding}
\usepackage{multirow}
\usepackage{makecell}
\usepackage[linesnumbered,ruled,vlined]{algorithm2e}
\usepackage{amsmath,amssymb,amsfonts}
\usepackage[table,xcdraw]{xcolor}
\usepackage{cite}
\usepackage{mathptmx} 
\usepackage[colorlinks,linkcolor=blue]{hyperref}
\usepackage{cleveref}
\usepackage{tabularx}
\usepackage[caption=true,font=normalsize,labelfont=sf,textfont=sf]{subfig}
\usepackage{textcomp}
\usepackage{stfloats}
\usepackage{tabularx}

\hypersetup{
    colorlinks=true, 
    linkcolor=blue,  
    filecolor=blue,  
    urlcolor=black,   
    citecolor=blue,  
    pdfborder={0 0 0} 
}

\Crefname{figure}{Fig.}{Figs.}
\Crefname{equation}{Eq.}{Eqs.}

\def\eg{\emph{e.g.}}

\def\ie{\emph{i.e.}}

\title{\LARGE \bf Enhancing Large Vision Model in Street Scene Semantic Understanding through Leveraging Posterior Optimization Trajectory}

\author{Wei-Bin Kou$^{1}$, Qingfeng Lin$^{1}$, Ming Tang$^{3}$, Jingreng Lei$^{1}$, Shuai Wang$^{4}$, \\Rongguang Ye$^{3}$, Guangxu Zhu$^{2,*}$, and Yik-Chung Wu$^{1,*}$
\thanks{$^{*}$Corresponding author: Guangxu Zhu (gxzhu@sribd.cn) and Yik-Chung Wu (ycwu@eee.hku.hk).}
\thanks{$^{1}$Department of Electrical and Electronic Engineering, The University of Hong Kong, Hong Kong 999077, China.}%
\thanks{$^{2}$Shenzhen Research Institute of Big Data, Shenzhen, China.}%
\thanks{$^{3}$Department of Computer Science and Engineering, Southern University of Science and Technology, Shenzhen, China.}%
\thanks{$^{4}$Shenzhen Institute of Advanced Technology, Chinese Academy of Sciences, Shenzhen, China.}%
}

\begin{document}

\maketitle
\thispagestyle{empty}
\pagestyle{empty}

\begin{abstract}
To improve the generalization of the autonomous driving (AD) perception model, vehicles need to update the model over time based on the continuously collected data. As time progresses, the amount of data fitted by the AD model expands, which helps to improve the AD model generalization substantially. However, such ever-expanding data is a double-edged sword for the AD model. Specifically, as the fitted data volume grows to exceed the the AD model's fitting capacities, the AD model is prone to under-fitting. To address this issue, we propose to use a pretrained Large Vision Models (LVMs) as backbone coupled with downstream perception head to understand AD semantic information. This design can not only surmount the aforementioned under-fitting problem due to LVMs' powerful fitting capabilities, but also enhance the perception generalization thanks to LVMs' vast and diverse training data. On the other hand, to mitigate vehicles' computational burden of training the perception head while running LVM backbone, we introduce a \underline{P}osterior \underline{O}ptimization \underline{T}rajectory (POT)-\underline{Gui}ded optimization scheme (POTGui) to accelerate the convergence. Concretely, we propose a POT \underline{Gen}erator (POTGen) to generate posterior (future) optimization direction in advance to guide the current optimization iteration, through which the model can generally converge within 10 epochs. Extensive experiments demonstrate that the proposed method improves the performance by over 66.48\% and converges faster over 6 times, compared to the existing state-of-the-art approach.
\end{abstract}

\section{INTRODUCTION}
\label{sec:intro}
Street scene semantic understanding in Autonomous Driving (AD) is a highly crucial but complex task \cite{10416354,10342110,10342254,10341639,kou2024fedrc,kou2024fast,kou2024adverse}. One major challenge in developing an effective and robust AD system is the poor generalization of the AD model due to the significant data heterogeneity \cite{10160999,9811702} in domain-shift setting, which is frequently observed in AD scenarios. For example, an AD vehicle transitioning into an unfamiliar environment may experience a notable decline in performance compared to operations within known settings. To improve the model generalization, vehicles need to train the model over time using the continuously collected data. As time progresses, the amount of the data fitted by the AD model continually expands, which helps to lead to a consistent and substantial improvement in the AD model generalization. However, such ever-increasing data is a double-edged sword for the AD model. Specifically, as the volume of dynamically collected data grows to exceed AD model's fitting capacity, the risk of under-fitting of the AD model increases, leading to the reduced accuracy in performance.

\begin{figure*}[tp]
\centering
\vspace{-0.2cm}
\includegraphics[width=\linewidth]{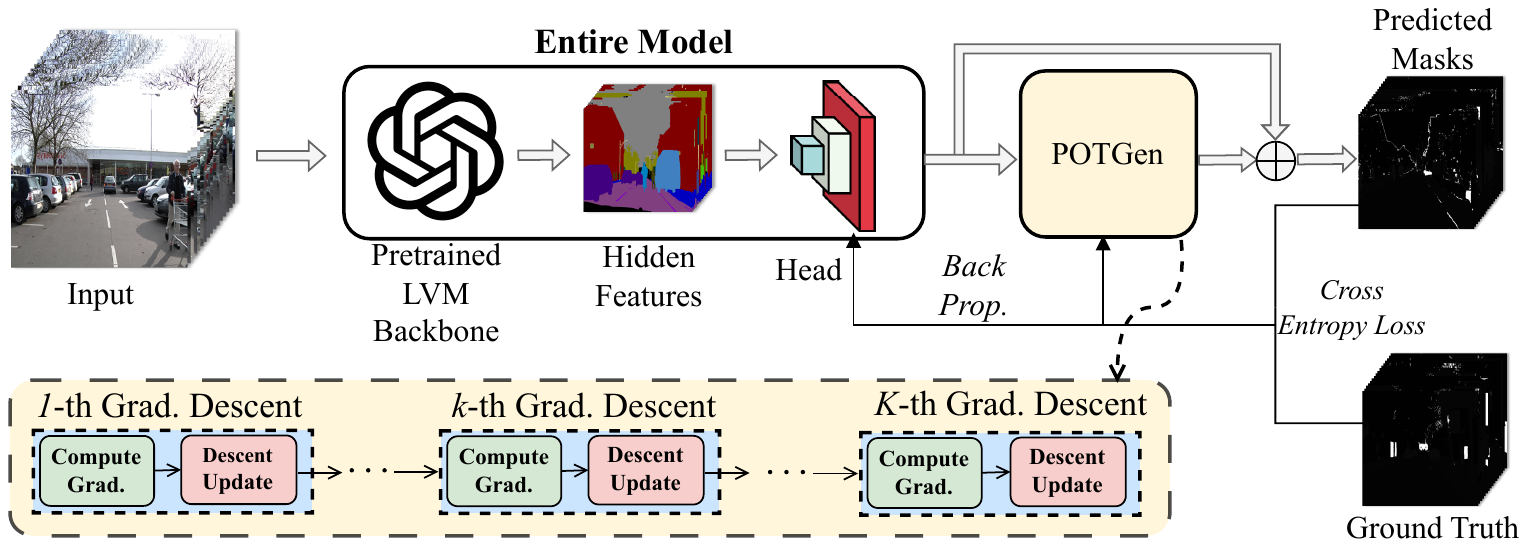}
\vspace{-7.9cm}
\caption{Illustration of the proposed POT-guided LVM-driven street scene semantic understanding method.}
\label{fig:overview}
\vspace{-0.5cm}
\end{figure*}

To tackle such under-fitting problem, we propose to use LVMs as backbone to fit the ever-expanding data in a zero-shot manner because: \textbf{(I) Depth and Width:} LVMs have more layers (depth) and more neurons per layer (width). This allows them to extract a hierarchy of data representations, from simple to complex, and capture more intricate patterns of the data. \textbf{(II) Vast Learnable Parameters:} The large number of parameters in LVMs allows them to effectively capture and learn the underlying distribution of the data, especially for complex tasks like semantic understanding in the context of AD. \textbf{(III) Attention Mechanism:} LVMs generally incorporate attention mechanisms, which allow the model to focus on different parts of the input when generating each part of the output. This leads to a more context-aware representation of the data and results in more meaningful feature extraction. On top of the LVM backbone, we propose to train a downstream perception head based on vehicle's onboard dataset. This perception head complements the LVM to understand the semantic information of driving surroundings. 

However, owing to vehicles' limitations of computational resource, training the proposed perception head while running the LVM backbone is time-consuming. To tackle this issue, we propose a Posterior Optimization Trajectory (POT)-Guided optimization scheme (POTGui) to accelerate the training. Specifically, a POT Generator (POTGen) is designated to generate the posterior (future) optimization direction in advance to guide the current optimization iteration. This innovative approach enables the model to converge typically within 10 epochs. This rapid convergence is helpful to mitigate the computation burden and is crucial for AD vehicles, where computational efficiency and timely decision-making are paramount. On the other hand, POTGui also poses a form of regularization, which inherently constrains the learning process, potentially improving the model performance and model generalization to new data.

Our main contributions are highlighted as follows:
\begin{itemize}
    \item This work proposes to leverage LVM backbone coupled with subsequent perception head to understand semantic information of driving environment. LVM can not only overcome the under-fitting problem, but also improve the AD model generalization across diverse scenarios. 
    \item To overcome vehicles' computational limitations of training the perception head while running LVM backbone, we propose POTGui to accelerate the perception head convergence. In addition, POTGui can also improve the model performance. 
    \item Extensive experiments demonstrate that the proposed method improves the model performance by 66.48\% and converges faster over 6 times, relative to existing state-of-the-art (SOTA) benchmarks.
\end{itemize}

\section{Related Work}
\label{related_work}
\subsection{Large Vision Models (LVMs)}
Recently, Large Language Models (LLMs) \cite{10610948,10611614} have achieved great success in the natural language processing (NLP) field in terms of various scenarios, such as language understanding and generation~\cite{Zhu2023mini}, performing user intent understanding~\cite{ouyang2022training}, knowledge utilization~\cite{jiang2023structgpt} and complex reasoning~\cite{wei2022chain} in a zero-shot/few-shot setting. Inspired by the achievements of pre-trained LLMs in NLP field, researchers have turned their attention to exploring pre-trained LVMs in computer vision. These models, pre-trained on extensive image datasets, hold the ability to decipher image content and distill rich semantic information. Prominent examples of such pre-trained LVMs include \cite{huang2023visual}. By learning representations and features from a significant volume of data, these models enhance the ability of computers to comprehend and analyze images more effectively, facilitating a range of diverse downstream applications. In this paper, we propose to use LVMs to tackle the under-fitting problem towards vehicles' ever-expanding fitted data thanks to their exceptional fitting capabilities.

\subsection{Autonomous Driving Semantic Understanding}
Semantic understanding is a field within computer vision and robotics focused on enabling machines to interpret and understand the semantic information of vehicles' surroundings, typically through various forms of sensory data such as images and lidars. This capability is crucial for AD \cite{10388394,9968085,10049523,10168231,10161421} to understand the layout of the street scene, including the road, pedestrian, sidewalks, buildings, and other static and dynamic elements. Modern semantic understanding heavily relies on machine learning (ML), particularly deep learning (DL) techniques. Initially, Fully Convolutional Networks (FCNs)-based models significantly improve the performance of this task \cite{yang2022deaot,zhou2022rethinking}. In recent years, Transformer-based approaches \cite{xie2021segformer} have also been proposed for semantic segmentation. Recently, Bird's Eye View (BEV) \cite{9697426} technique is widely adopted for road scene understanding. 
Moreover, some works have been done to improve the AD model generalization by adopting Federated Learning (FL) \cite{feddrive2022,10342134}.
In this paper, we propose to use LVMs backbone coupled with downstream perception head to understand semantic information of vehicles' surroundings. In addition, we also propose POTGui to accelerate the model convergence and model performance. 

\section{Methodology}
\label{methodology}
\subsection{LVM-Driven Model Architecture}
Pretrained ImageGPT \cite{chen2020generative}, often abbreviated as iGPT, is an outstanding representative of LVMs and is selected as the LVM backbone (with parameters $\omega_{lvm}$), which is utilized to extract hidden features $\mathcal{F}_{h}^{(j)}$ of $j$-th mini-batch data $\mathcal{D}_v^{(j)}$ from training dataset $\mathcal{D}_v$ in a zero-shot fashion, \ie, 
\begin{align}
\mathcal{F}_{h}^{(j)} = \omega_{lvm}(\mathcal{D}_v^{(j)}).
\label{Eq:lvm_forward}
\end{align}
$\mathcal{F}_{h}^{(j)}$ is high-dimensional vectors that represents the model's understanding of the image content, and can be used as input to train the downstream perception head. 

Once $\mathcal{F}_{h}^{(j)}$ has been extracted by the LVM backbone, it is then transmitted to downstream perception head (with parameters $\omega_{su}$) as input. Then vehicle can train the perception head based on $\mathcal{F}_{h}^{(j)}$ and ground truth $P_Y^{(j)}$. Specifically, the cross entropy loss of $j$-th mini-batch is defined as $L_{CE}(P_Y^{(j)}, \mathcal{O}^{(j)})$ to minimize the distance between the ground truth $P_Y^{(j)}$ and the predicted logits $\mathcal{O}^{(j)}$. The training of the perception head is given by 
\begin{align}
   \mathcal{O}^{(j)} &= \omega_{su}(\mathcal{F}_{h}^{(j)}), 
   \label{Eq:perc_forward}
   \\
\mathop{\mathrm{min}}_{\mathbf{\omega}_{su}}~ L(\mathcal{\omega}_{su}) &=
\frac{1}{|\mathcal{D}_{v}|}
\sum\nolimits_{\mathcal{D}_{v}^{(j)} \in \mathcal{D}_{v}} L_{CE}(P_Y^{(j)}, \mathcal{O}^{(j)}).
\label{Eq:opt_loss}
\end{align}
Specifically, for each iteration, when the cross entropy loss $L_{CE}(P_Y^{(j)}, \mathcal{O}^{(j)})$ over $\mathcal{D}_v^{(j)}$ is calculated, $\mathcal{\omega}_{su}$ is optimized by back propagation. 

In addition, ASSP \cite{chen2017deeplab} is proposed to serve as the architecture of the downstream perception head, thanks to its capability to capture multi-scale context by aggregating features from various receptive field sizes. In summary, the LVM backbone coupled with the downstream perception head, serving as the \textbf{Entire Model} (denoted as model hereafter), is illustrated in \Cref{fig:overview}.

\begin{figure}[tp]
\vspace{-0.4cm}
\hspace{-0.58cm}
\includegraphics[width=1.1\linewidth, height=1.4\linewidth]{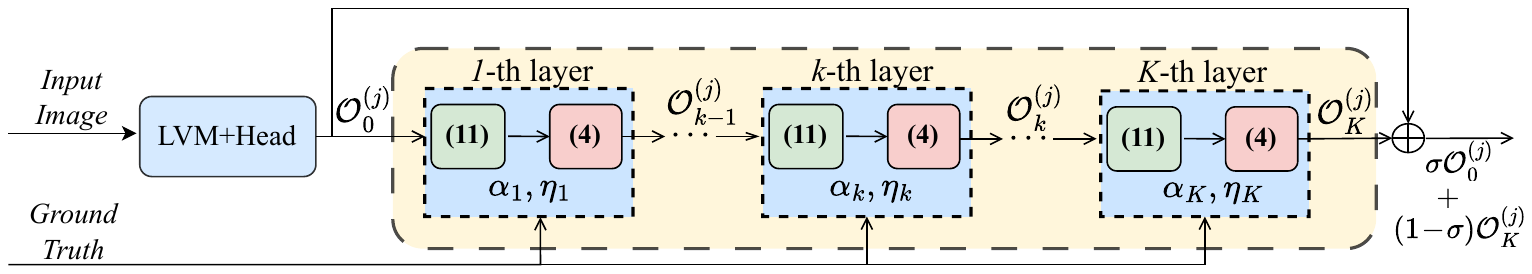}
\vspace{-9.3cm}
\caption{Illustration of the proposed POTGui optimization scheme. (4), (12) represent equation number.}
\label{Fig.RDU}
\end{figure}

\subsection{POT Generator (POTGen)}
To generate POT for current training iteration in advance, inspired by \cite{10529194,lin2023communication}, we propose to treat each optimization iteration as one layer. Based on this, cross entropy loss $L_{CE}(\cdot, \cdot)$ can be unfolded into $K$ layers at the end of forward propagation (illustrated in \Cref{Fig.RDU}). Specifically, for layer $k,\ \text{where}\ k \in \{1, 2, \cdots, K\}$, the predicted logits associated with $\mathcal{D}_v^{(j)}$ is updated as follow:
\begin{align}
    \mathcal{O}^{(j)}_k = \mathcal{O}^{(j)}_{k-1}-\eta_k \alpha_k \nabla_\mathcal{O} L_{CE}(P_Y^{(j)}, \mathcal{O}^{(j)}_{k-1}),
    \label{Eq:RDU_update}
\end{align}
where $\alpha_{k}$ and $\eta_{k}$ are two learnable parameters to control the update step size collectively. $\nabla_\mathcal{O} L_{CE}(P_Y^{(j)}, \mathcal{O}_{k-1}^{(j)})$ is the gradients of $L_{CE}(\cdot, \cdot)$ relative to $(k\!-\!1)$-th layer's output $\mathcal{O}_{k-1}^{(j)}$.

To calculate $\nabla_\mathcal{O} L_{CE}(P_Y^{(j)}, \mathcal{O}_{k-1}^{(j)})$, we firstly formulate the cross entropy loss $L_{CE}(P_Y^{(j)}, \mathcal{O}_{k-1}^{(j)})$ as follow: 
\begin{align}
\!L_{CE}(P_Y^{(j)}, \mathcal{O}_{k-1}^{(j)}) &=
L_{CE}(P_Y^{(j)}, P_{X, k-1}^{(j)}) \nonumber \\ &= -\frac{1}{N}\sum\nolimits_{i=1}^{N}\sum\nolimits_{c=1}^C P_Y^{(j, i, c)} \log(P_{X, k-1}^{(j, i, c)}),
\end{align}
where $N = |\mathcal{D}_v^{(j)}|$ (\ie, batch size), $C$ is the number of semantic classes, $P_Y^{(j, i, c)}$ is the ground truth of class $c$ of the $i$-th image in $\mathcal{D}_v^{(j)}$ (denoted as $\mathcal{D}_v^{(j, i)}$), $P_{X, k-1}^{(j, i, c)}$ represents the predicted probability of class $c$ for $\mathcal{D}_v^{(j, i)}$. Generally, $P_{X, k-1}^{(j, i, c)}$ is the output of softmax function of the predicted logits, \ie,
\begin{align}
P_{X, k-1}^{(j, i, c)} = {exp({\mathcal{O}_{k-1}^{(j, i, c)}}})/\ {\sum\nolimits_{c=1}^C exp({\mathcal{O}_{k-1}^{(j, i, c)}}}),
\end{align}
where $\mathcal{O}_{k-1}^{(j, i, c)}$ is the predicted logits for class $c$ for $\mathcal{D}_v^{(j, i)}$. 

\begin{figure}[tp]
\centering
\subfloat[POTGui Optimization]{\includegraphics[width=0.5\linewidth, height=0.4\linewidth]{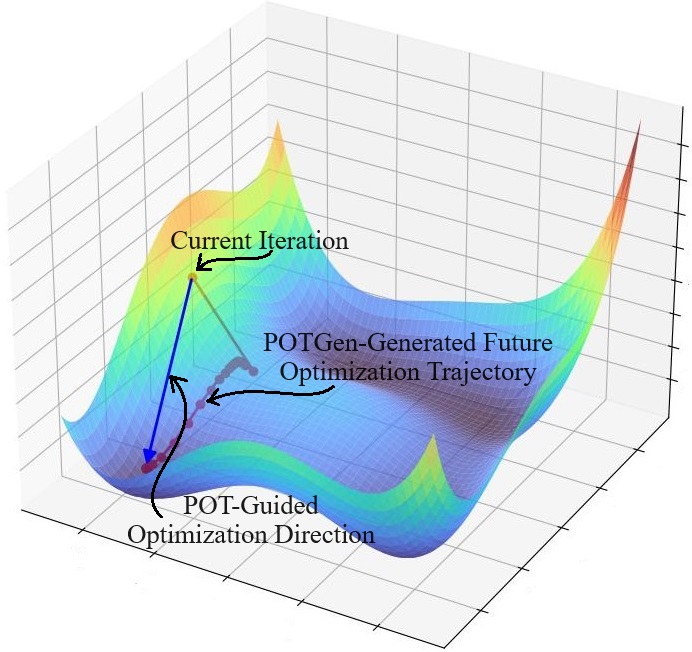}%
\label{Fig:POTGen_process}}
\subfloat[mIoU Comparison]{\includegraphics[width=0.5\linewidth, height=0.4\linewidth]{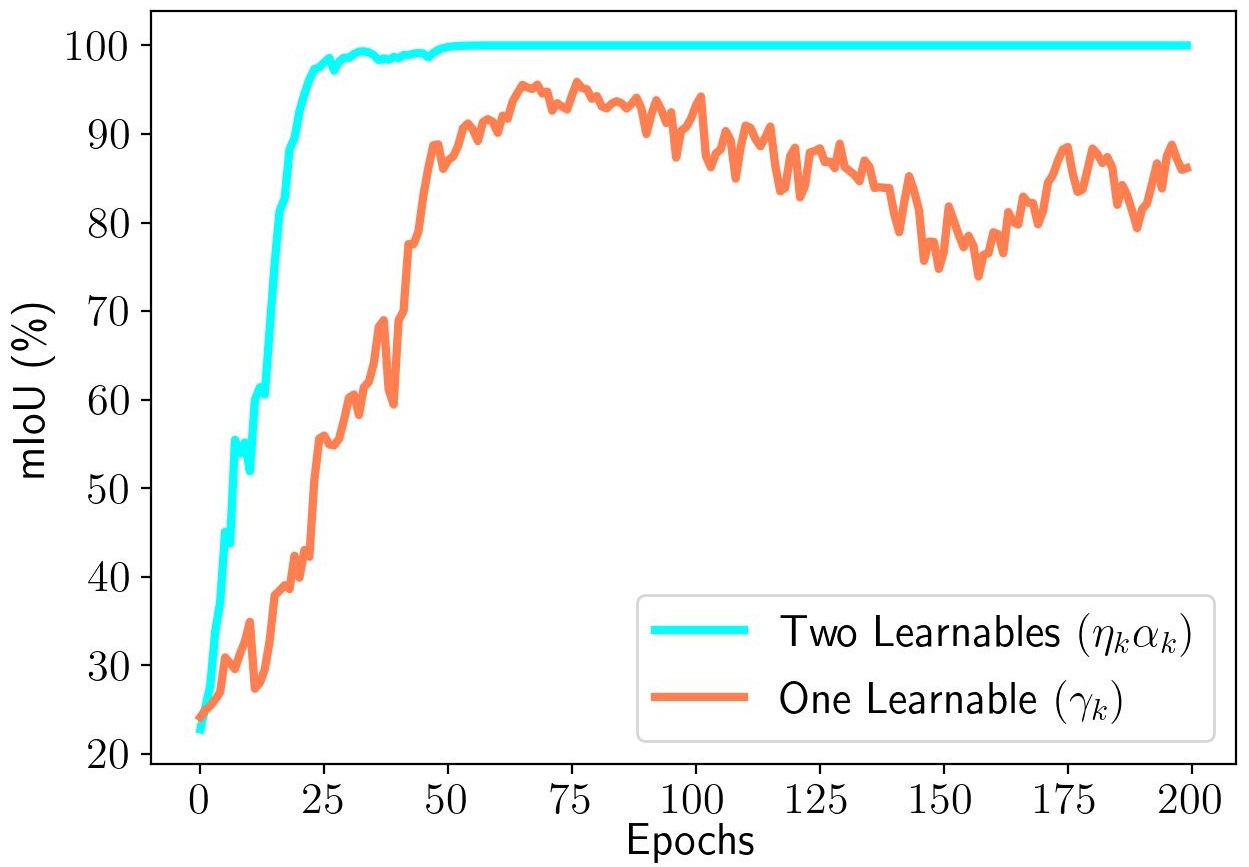}%
\label{Fig:learnalbe2_vs_learnable1}}
\caption{In-depth exploration of POTGui optimization scheme.}
\label{Fig:in_depth_POTGui}
\end{figure}

Based on above formulation of $L_{CE}(P_Y^{(j)}, \mathcal{O}_{k-1}^{(j)})$, we calculate the gradients $\nabla_\mathcal{O} L_{CE}(P_Y^{(j)}, \mathcal{O}_{k-1}^{(j)})$ by applying the chain rule:
\begin{align}
\frac{\partial L_{CE}(P_Y^{(j)}, \mathcal{O}_{k-1}^{(j)})}{\partial \mathcal{O}_{k-1}^{(j, i, c)}} = -\frac{1}{N}\sum_{i=1}^{N}\sum_{c=1}^C \frac{\partial L_{CE}(P_Y^{(j)}, \mathcal{O}_{k-1}^{(j)})}{\partial P_{X, k-1}^{(j, i, c)}} \frac{\partial P_{X, k-1}^{(j, i, c)}}{\partial \mathcal{O}_{k-1}^{(j, i, c)}},
\label{Eq:LCE_grad}
\end{align}
where
\begin{align}
\frac{\partial L_{CE}(P_Y^{(j)}, \mathcal{O}_{k-1}^{(j)})}{\partial P_{X, k-1}^{(j, i, c)}}
&= \frac{P_Y^{(j, i, c)}}{P_{X, k-1}^{(j, i, c)}},
\label{Eq:L_to_PX}
\\
\frac{\partial P_{X, k-1}^{(j, i, c)}}{\partial \mathcal{O}_{k-1}^{(j, i, c)}}
&= P_{X, k-1}^{(j, i, c)} (1 - P_{X, k-1}^{(j, i, c)}).
\label{Eq:PX_to_logits}
\end{align}
Substituting \Cref{Eq:L_to_PX,Eq:PX_to_logits} into \Cref{Eq:LCE_grad}, we get:
\begin{align}
\!\frac{\partial L_{CE}(P_Y^{(j)}\!, \mathcal{O}_{k-1}^{(j)})}{\partial \mathcal{O}_{k-1}^{(j, i, c)}}\!=\!-\frac{1}{N} \sum_{i=1}^{N}(P_Y^{(j, i, c)} \!-\! \sum_{t=1}^C P_{X, k-1}^{(j, i, t)} P_Y^{(j, i, t)}).
\end{align}
Since $\sum_{t=1}^C P_{X, k-1}^{(j, i, t)} P_Y^{(j, i, t)} = P_{X, k-1}^{(j, i, c)}$ ($P_Y^{(j, i, t)} = 0\ \text{if}\ t \neq c; P_Y^{(j, i, t)} = 1\ \text{if}\ t = c$), therefore, the gradients $\nabla_\mathcal{O} L_{CE}(P_Y^{(j)}, \mathcal{O}_{k-1}^{(j)})$ can be simplified to:
\begin{align}
\frac{\partial L_{CE}(P_Y^{(j)}, \mathcal{O}_{k-1}^{(j)})}{\partial \mathcal{O}_{k-1}^{(j, i, c)}} = \frac{1}{N} \sum\nolimits_{i=1}^{N}(P_{X, k-1}^{(j, i, c)} - P_Y^{(j, i, c)}).
\label{Eq:grad_of_LCE}
\end{align}

After the updates by $K$ layers, the output optimized logits of POTGen is:
\begin{align}
   \mathcal{O}^{(j)}_{POT} = \mathcal{O}^{(j)}_{K} = \mathcal{O}^{(j)}_0 - \sum\nolimits_{k=1}^K \eta_k \alpha_k \nabla_\mathcal{O} L_{CE}(P_Y^{(j)}, \mathcal{O}^{(j)}_{k-1}),
   \label{Eq:potgen}
\end{align}
where $\mathcal{O}^{(j)}_0$ (equal to $\mathcal{O}^{(j)}$) represents the original predicted logits of the model. We can observe that $\mathcal{O}^{(j)}_{POT}$ contains future steps' optimization trajectory with respect to current iteration.

\subsection{POT-Guided Optimization (POTGui)}
The output of POTGen $\mathcal{O}^{(j)}_{POT}$ is then used to guide the current optimization iteration. Specifically, $\mathcal{O}^{(j)}_{POT}$ is added to the original logits $\mathcal{O}^{(j)}$ of the model in a weighted form, \ie,
\begin{align}
    \mathcal{O}^{(j)}_{POTGui} = \sigma\mathcal{O}^{(j)} + (1 - \sigma) \mathcal{O}^{(j)}_{POT}, ~~\sigma \in [0, 1].
\end{align}
Since $\mathcal{O}^{(j)}_{POT}$ results from multiple gradient descent updates specifically tailored to optimize $L_{CE}(\cdot, \cdot)$, its weighted addition to the original logits $\mathcal{O}^{(j)}$ can guide the overall model predictions towards these optimized outcomes. This process is demonstrated in \Cref{Fig:POTGen_process}.

Then we use the summation logits $\mathcal{O}^{(j)}_{POTGui}$ to take place of original logits $\mathcal{O}^{(j)}$ in \Cref{Eq:opt_loss} to calculate the loss to optimize the model by back propagation, \ie,
\begin{align}
\mathop{\mathrm{min}}_{\mathbf{\omega}_{su}}~ L(\mathcal{\omega}_{su}) =
\frac{1}{|\mathcal{D}_{v}|}
\sum\nolimits_{\mathcal{D}_{v}^{(j)} \in \mathcal{D}_v} L_{CE}(P_Y^{(j)}, \mathcal{O}_{POTGui}^{(j)}).
\label{Eq:opt_loss_potgui}
\end{align}
In addition, the learnable parameters $\alpha_k$, $\eta_k,\ \text{where}\ k \in \{1, 2, \cdots K\}$ are also optimized by back propagation along with the model parameters $\omega_{su}$.

We try to use one learnable parameter (\eg, $\gamma_k$) instead of two parameters $\eta_k \alpha_k$ in each layer. \Cref{Fig:learnalbe2_vs_learnable1} compares the performance of these two cases and indicates that the case with two learnable parameters outperforms the case with one learnable parameters. This is because that two learnable parameters have more powerful fitting capabilities. 
In conclusion, the proposed POTGui is outlined in Algo. \ref{Algo:DUN}.

\begin{algorithm}[tp]
\caption{POTGui Optimization Scheme}
\label{Algo:DUN}
\SetAlgoLined
\KwIn{$\mathcal{D}_v$ (Training Dataset), $P_Y$ (One-hot ground truth), $K$ (Layer number), $epochs$ (Training epochs), $\sigma$ (Addition weight)}
\KwOut{Model $model$, Learnable variables $\alpha_{k}, \eta_{k}, \ k \in \{1, 2, \cdots K\}$}
Initialize $model \gets \mathcal{W}_0$, $\alpha_{k}, \eta_{k} \gets \alpha_0, \eta_0, \ k \in \{1, 2, \cdots K\}$ \\
\For{$epoch\ e \gets 1$ \KwTo $epochs$}{ 
\For {$\mathcal{D}_v^{(j)}  \in \mathcal{D}_v$}{
$\mathcal{O}_0^{(j)} \gets \mathcal{O}^{(j)} \gets model(\mathcal{D}_v^{(j)})$ \\
\For{$layer\ k \gets 1$ \KwTo $K$}{ 
    $\nabla_\mathcal{O} L_{CE}(P_Y^{(j)}, \mathcal{O}^{(j)}_{k-1}) \gets \Cref{Eq:grad_of_LCE}$
    $\mathcal{O}^{(j)}_k \gets \Cref{Eq:RDU_update}$
    \tcp{Update rule}
}
$\mathcal{O}^{(j)}_{POTGui}\!\gets\!\sigma\mathcal{O}^{(j)}\! +\! (1\!-\!\sigma)\mathcal{O}^{(j)}_{K}$ 
$\mathcal{L} = L_{CE} (P_Y^{(j)}, \mathcal{O}^{(j)}_{POTGui})$ 
$model, \alpha_{k}, \eta_{k},\ k \in \{1, 2, \cdots K\} \gets \mathcal{L}.Backward()$ 
}
}
\end{algorithm}

\section{Experiments}
\label{experiments}
In this section, we carry out comprehensive experiments to verify the proposed method in the context of AD. Hereafter, we denote LVM+Head as LVM and LVM+Head+POTGui as LVM+POTGui for short. 

\subsection{Datasets, Evaluation Metrics and Implementation}
\subsubsection{Datasets}
The \textbf{Cityscapes} dataset \cite{Cordts2016Cityscapes}
consists of 2,975 training and 500 validation images with ground truth. Training dataset includes pixel-level label of 19 classes, including vehicles, pedestrians and so forth. The \textbf{CamVid} dataset \cite{brostow2008segmentation} totally includes 701 images with pixel-level label of 11 classes. We randomly select 600 images to form training dataset, and the remaining 101 images are served as test dataset. In addition, we will also conduct real driving test on Apolloscapes dataset \cite{wang2019apolloscape} and CARLA\_ADV dataset captured from CARLA \cite{dosovitskiy2017carla} simulator under various weather conditions, such as foggy, rainy, cloudy, etc.

\subsubsection{Evaluation Metrics}
We evaluate the proposed method on street scene semantic understanding task by employing four widely used metrics: Mean Intersection over Union (\textbf{mIoU}), which measures the overlap between predicted mask and ground truth; Mean Precision (\textbf{mPre}), which assesses the accuracy of positive predictions; Mean Recall (\textbf{mRec}), evaluating how well the model identifies all relevant instances; and Mean F1 (\textbf{mF1}), which provides a balance between precision and recall. 

\subsubsection{Implementation Details}
The deep learning model was implemented using the Pytorch framework, with experiments conducted on two NVIDIA GeForce 4090 GPUs. We select the Adam optimizer for training, configuring it with Betas values of 0.9 and 0.999, and set the weight decay at 1e-4. The training was executed with a batch size of 8 and a learning rate of 3e-4. $\sigma$ in POTGui is set to 0.5.

\subsection{Main Results and Empirical Analysis}
In this section, we will present experimental results and conduct empirical analysis in following aspects:

\begin{table*}[tp]
\centering
\setlength{\tabcolsep}{11.9pt}
\caption{Performance comparison of various LVM hidden features on Cityscapes and CamVid dataset}
\begin{tabularx}{\linewidth}{ccccclcccc}
\hline
\multirow{2}{*}{Feature Layer(s)} & \multicolumn{4}{c}{Cityscapes Dataset (19 Semantic Classes) (\%)}          &  & \multicolumn{4}{c}{CamVid Dataset (11 Semantic Classes) (\%)}              \\ \cline{2-5} \cline{7-10} 
                                  & mIoU           & mF1            & mPrecision     & mRecall        &  & mIoU           & mF1            & mPrecision     & mRecall        \\ \hline
iGPT\_All\_Avg                    & 43.70          & 53.45          & 54.16          & 54.71          &  & 48.59          & 60.37          & 69.40          & 56.67          \\
iGPT\_Last                        & 39.15          & 49.37          & 51.94          & 49.92          &  & 45.07          & 56.18          & 64.57          & 53.69          \\
iGPT\_Middle\_1                   & 43.49          & 53.24          & 54.41          & 53.87          &  & 48.28          & 60.00          & 69.03          & 56.37          \\
iGPT\_Middle\_4\_Avg              & 43.22          & 53.02          & 55.14          & 54.16          &  & 48.84          & 60.59          & 69.99          & 56.74          \\
iGPT\_Middle\_4                   & \textbf{45.81} & \textbf{55.15} & \textbf{56.18} & 56.00 &  & \textbf{49.29} & \textbf{60.90} & \textbf{70.52} & 57.27          \\
iGPT\_All                                      & 44.76   & 54.34   & 55.28 & \textbf{56.38}             &  & 48.98          & 59.51          & 70.12          & \textbf{57.47} \\ \hline
\end{tabularx}
\label{Tab:iGPT_hidden_feats}
\vspace{-0.3cm}
\end{table*}

\subsubsection{Hidden Feature Selection from LVM}
In this part, we intend to pinpoint the most potent features out of following six distinct cases: I. Features extracted from the final layer (\ie, LVM\_Last); II. Features from the central layer (\ie, LVM\_Middle\_1); III. Averaging features across all layers (\ie, LVM\_ALL\_Avg); IV. Averaging features from the central four layers (\ie, LVM\_Middle\_4\_Avg); V. Features from the middle four layers (\ie, LVM\_Middle\_4); and VI. Features from all layers (\ie, LVM\_ALL). This thorough comparative study is designed to ascertain the best layer (or layers) tailored to the needs of AD semantic understanding application. \Cref{Tab:iGPT_hidden_feats} compares the performance of aforementioned cases quantitatively, and demonstrates that LVM\_Middle\_4 yields the highest scores across almost all evaluation metrics. Based on this, subsequent experiments are based on LVM\_Middle\_4.

\begin{table*}[tp]
\centering
\setlength{\tabcolsep}{11.5pt}
\caption{Average inference performance comparison of all semantic classes on both Cityscapes and CamVid dataset}
\begin{tabularx}{\linewidth}{ccccclcccc}
\hline
\multirow{2}{*}{Benchmarks} & \multicolumn{4}{c}{Cityscapes Dataset (19 Semantic Classes) (\%)}          &  & \multicolumn{4}{c}{CamVid Dataset (11 Semantic Classes) (\%)}              \\ \cline{2-5} \cline{7-10} 
                                  & mIoU           & mF1            & mPrecision     & mRecall                &  & mIoU           & mF1            & mPrecision     & mRecall        \\ \hline
BiSeNetV2 \cite{yu2021bisenet}           & 33.63          & 43.32          & 44.73          & 43.96           &  & 47.89          & 53.33          & 55.12          & 53.33          \\
SegNet \cite{badrinarayanan2017segnet}   & 43.14          & 52.87          & 53.47          & 53.54           &  & 46.60          & 50.18          & 49.42          & 51.26        \\
DeepLabv3+ \cite{chen2018encoderdecoder}                               & 69.04          & 75.95          & 75.29          & 77.57           &  & 69.46          & 77.58          & 81.10          & 76.19          \\
SegFormer \cite{xie2021segformer}                                & 39.37          & 46.23          & 43.60          & 50.07           &  & 34.23          & 38.86          & 37.26          & 41.04          \\
LVM \cite{chen2020generative}                                 & 45.81          & 55.15          & 56.18          & 56.00           &  & 49.29          & 60.90          & 70.52          & 57.27          \\
\textbf{LVM+POTGui (Ours)}                 & \textbf{99.99}          & \textbf{99.99}          & \textbf{99.99}          & \textbf{99.99}           &  & \textbf{82.06}          & \textbf{89.52}          & \textbf{97.70}          & \textbf{85.51}          \\
\hline
\end{tabularx}
\label{Tab:iGPT_tradi}
\vspace{-0.4cm}
\end{table*}

\begin{figure*}[!t]
\centering
\subfloat[mIoU on Cityscapes]{\includegraphics[width=0.25\linewidth]{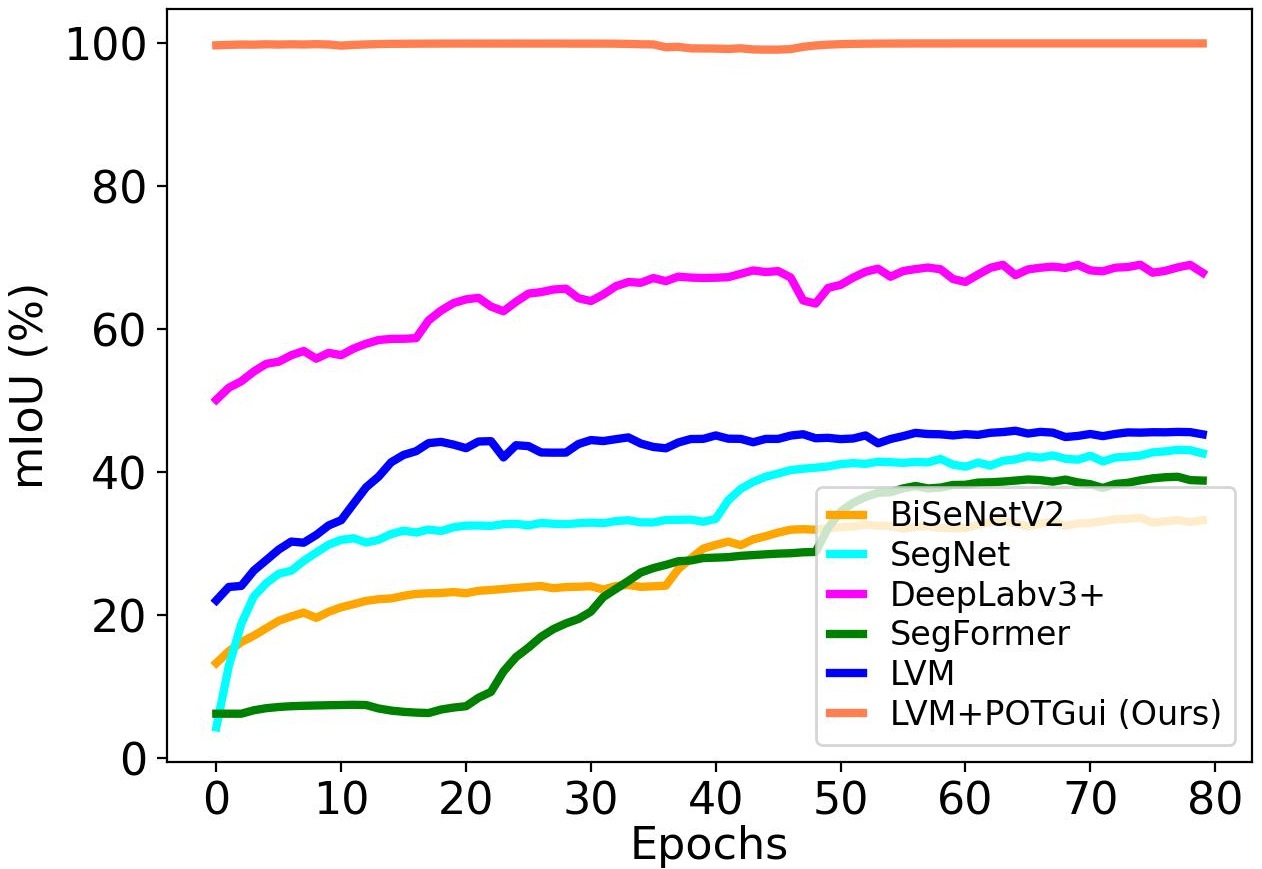}%
\label{Fig:iGPT_tradi_e}}
\subfloat[mF1 on Cityscapes]{\includegraphics[width=0.25\linewidth]{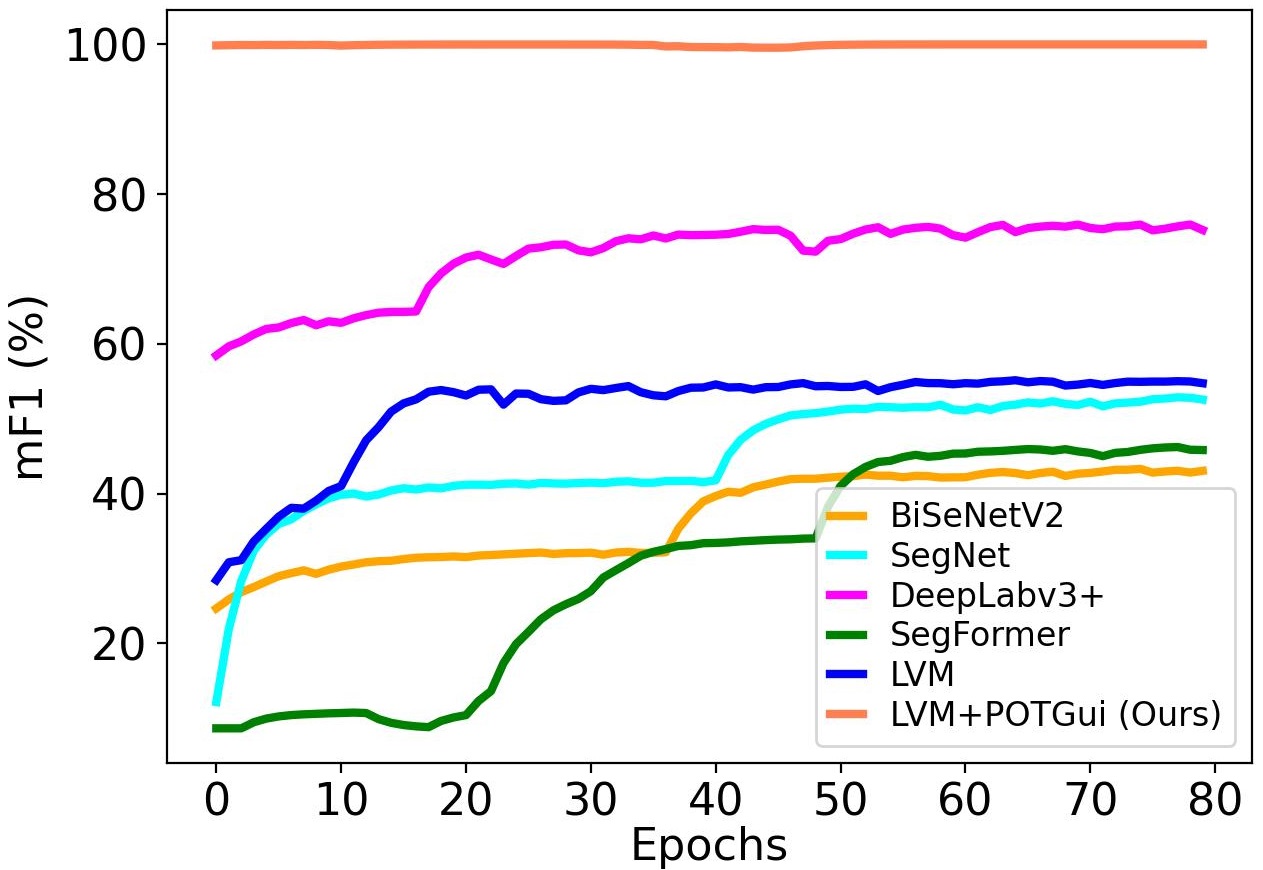}%
\label{Fig:iGPT_tradi_h}}
\subfloat[mIoU on CamVid]{\includegraphics[width=0.245\linewidth]{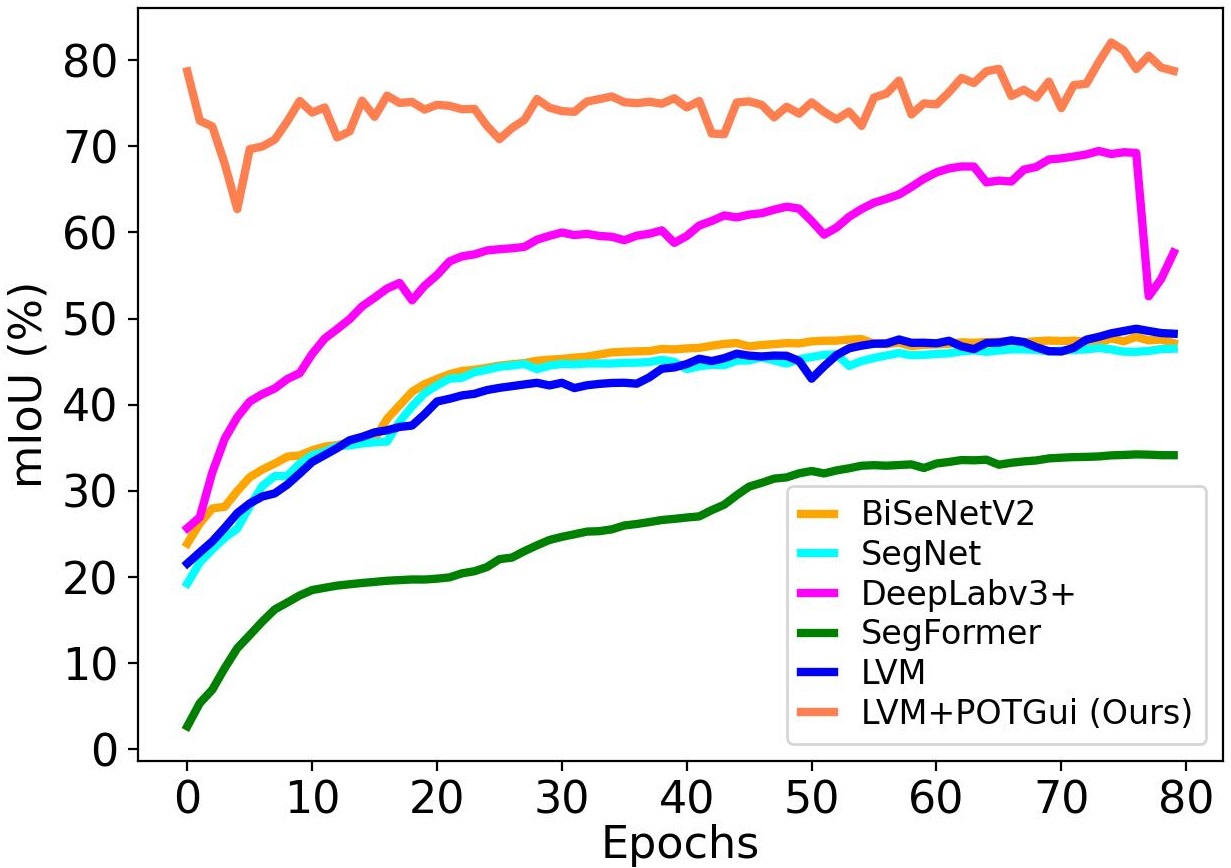}%
\label{Fig:iGPT_tradi_a}}
\subfloat[mF1 on CamVid]{\includegraphics[width=0.245\linewidth]{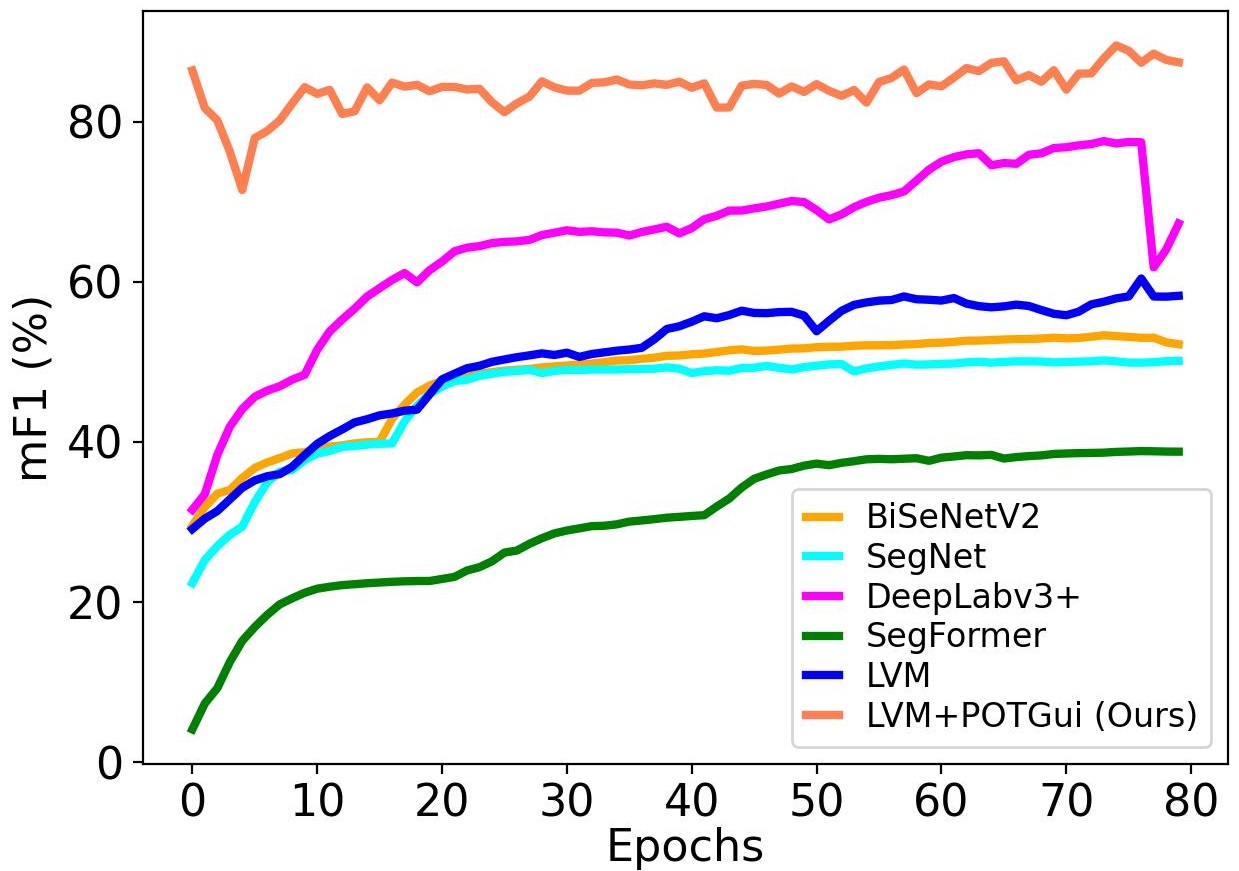}%
\label{Fig:iGPT_tradi_d}}
\caption{Illustration of performance of LVM+POTGui against existing state-of-the-art benchmarks on Cityscapes and CamVid.}
\label{Fig:iGPT_tradi}
\vspace{-0.4cm}
\end{figure*}

\begin{table*}[tp]
\setlength{\tabcolsep}{4.92pt}
\caption{Inference performance comparison for all semantic classes on CamVid dataset across all metrics}
\begin{tabularx}{\linewidth}{ccccccccccccc}
\hline
\multirow{2}{*}{Metric}& \multirow{2}{*}{Benchmarks}      & \multicolumn{11}{c}{CamVid Dataset (11 Semantic Classes) (\%)}                                                                     \\ \cline{3-13} 
                       &                                  & Sky   & Building & Pole  & Road  & Sidewalk & Tree  & Signsymbol & Fence & Car   & Pedestrian & Bicyclist \\ \hline
\multicolumn{1}{c|}{\multirow{6}{*}{IoU}} & \multicolumn{1}{c|}{BiSeNetV2 \cite{yu2021bisenet}}       & 92.78 & 84.60 & 0.00  & 96.37 & 84.67 & 79.58 & 19.54 & 0.00  & 81.20 & 0.00  & 0.00 \\
\multicolumn{1}{c|}{}  & \multicolumn{1}{c|}{SegNet \cite{badrinarayanan2017segnet}}                  & 93.93 & 83.09 & 0.00  & 95.98 & 82.60 & 79.15 & 0.00  & 0.00  & 79.18 & 0.00  & 0.00 \\
\multicolumn{1}{c|}{}  & \multicolumn{1}{c|}{DeepLabv3+ \cite{chen2018encoderdecoder}}                                              & 92.46 & 89.84 & 0.00  & 97.53 & 88.45 & 82.48 & 55.33 & 74.51 & 87.77 & 56.38 & 75.33 \\
\multicolumn{1}{c|}{}  & \multicolumn{1}{c|}{SegFormer \cite{xie2021segformer}}                                               & 92.47 & 78.00 & 0.00  & 92.23 & 68.43 & 73.16 & 0.00  & 0.00  & 64.00 & 0.00  & 0.00 \\
\multicolumn{1}{c|}{}  & \multicolumn{1}{c|}{LVM \cite{chen2020generative}}                                                & 85.31 & 74.60 & 0.05  & 89.15 & 63.81 & 64.60 & 23.80 & 34.30 & 50.80 & 4.36  & 44.80 \\
\multicolumn{1}{c|}{}  & \multicolumn{1}{c|}{\textbf{LVM+POTGui (Ours)}}                                & \textbf{98.43} & \textbf{96.55} & \textbf{85.27} & \textbf{98.33} & \textbf{94.77} & \textbf{95.64} & \textbf{82.12} & \textbf{93.24} & \textbf{94.75} & \textbf{85.82} & \textbf{91.59} \\ \hline
\multicolumn{1}{c|}{\multirow{6}{*}{F1}}  & \multicolumn{1}{c|}{BiSeNetV2 \cite{yu2021bisenet}}       & 96.25 & 91.66 & 0.00  & 98.15 & 91.70 & 88.63 & 31.73 & 0.00  & 89.62 & 0.00  & 0.00  \\
\multicolumn{1}{c|}{}  & \multicolumn{1}{c|}{SegNet \cite{badrinarayanan2017segnet}}                  & 96.87 & 90.77 & 0.00  & 97.95 & 90.46 & 88.36 & 0.00  & 0.00  & 88.37 & 0.00  & 0.00  \\
\multicolumn{1}{c|}{}  & \multicolumn{1}{c|}{DeepLabv3+ \cite{chen2018encoderdecoder}}                                              & 96.08 & 94.65 & 0.00  & 98.75 & 93.87 & 90.40 & 71.22 & 85.38 & 93.48 & 72.09 & 89.90  \\
\multicolumn{1}{c|}{}  & \multicolumn{1}{c|}{SegFormer \cite{xie2021segformer}}                                               & 96.09 & 87.64 & 0.00  & 95.96 & 81.25 & 84.49 & 0.00  & 0.00  & 78.05 & 0.00  & 0.00  \\
\multicolumn{1}{c|}{}  & \multicolumn{1}{c|}{LVM \cite{chen2020generative}}                                                & 92.07 & 85.43 & 0.09  & 94.27 & 77.91 & 78.47 & 38.44 & 51.03 & 67.37 & 8.15  & 61.83 \\
\multicolumn{1}{c|}{}  & \multicolumn{1}{c|}{\textbf{LVM+POTGui (Ours)}}                                & \textbf{99.21} & \textbf{98.24} & \textbf{92.02} & \textbf{99.16} & \textbf{97.31} & \textbf{97.77} & \textbf{90.14} & \textbf{96.49} & \textbf{97.30} & \textbf{92.34} & \textbf{95.60} \\ \hline
\multicolumn{1}{c|}{\multirow{6}{*}{Precision}} & \multicolumn{1}{c|}{BiSeNetV2 \cite{yu2021bisenet}} & 96.36 & 88.32 & 0.00  & 97.73 & 92.96 & 87.81 & 55.22 & 0.00  & 87.99 & 0.00  & 0.00   \\
\multicolumn{1}{c|}{}  & \multicolumn{1}{c|}{SegNet \cite{badrinarayanan2017segnet}}                  & 97.11 & 86.45 & 0.00  & 97.58 & 92.49 & 87.46 & 0.00  & 0.00  & 88.42 & 0.00  & 0.00   \\
\multicolumn{1}{c|}{}  & \multicolumn{1}{c|}{DeepLabv3+ \cite{chen2018encoderdecoder}}                                              & 96.50 & 93.40 & 0.00  & 98.68 & 94.52 & 90.48 & 96.64 & 90.14 & 94.38 & 78.11 & 89.59  \\
\multicolumn{1}{c|}{}  & \multicolumn{1}{c|}{SegFormer \cite{xie2021segformer}}                                               & 96.19 & 83.63 & 0.00  & 95.48 & 83.13 & 84.43 & 0.00  & 0.00  & 78.83 & 0.00  & 0.00  \\
\multicolumn{1}{c|}{}  & \multicolumn{1}{c|}{LVM \cite{chen2020generative}}                                                & 93.25 & 83.41 & 10.94 & 93.75 & 84.20 & 85.07 & 96.84 & 77.53 & 76.84 & 73.18 & 88.30 \\
\multicolumn{1}{c|}{}  & \multicolumn{1}{c|}{\textbf{LVM+POTGui (Ours)}}                                & \textbf{99.35} & \textbf{97.64} & \textbf{100.00}& \textbf{98.70} & \textbf{99.10} & \textbf{98.67} & \textbf{100.00}& \textbf{99.86} & \textbf{99.37} & \textbf{100.00}& \textbf{99.99} \\ \hline
\multicolumn{1}{c|}{\multirow{6}{*}{Recall}} & \multicolumn{1}{c|}{BiSeNetV2 \cite{yu2021bisenet}}    & 97.21 & 96.86 & 0.00 & 98.84 & 92.05  & 92.10 & 23.49 & 0.00  & 94.25 & 0.00  & 0.00  \\
\multicolumn{1}{c|}{}  & \multicolumn{1}{c|}{SegNet \cite{badrinarayanan2017segnet}}                  & 98.02 & 96.95 & 0.00 & 98.79 & 91.59  & 91.61 & 0.00  & 0.00  & 90.96 & 0.00  & 0.00 \\
\multicolumn{1}{c|}{}  & \multicolumn{1}{c|}{DeepLabv3+ \cite{chen2018encoderdecoder}}                                              & 97.08 & 97.04 & 0.00 & 99.11 & 94.41  & 91.90 & 63.96 & 85.72 & 93.82 & 70.60 & 88.05  \\
\multicolumn{1}{c|}{}  & \multicolumn{1}{c|}{SegFormer \cite{xie2021segformer}}                                               & 97.95 & 97.10 & 0.00 & 99.76 & 81.76  & 88.01 & 0.00  & 0.00  & 81.54 & 0.00  & 0.00   \\
\multicolumn{1}{c|}{}  & \multicolumn{1}{c|}{LVM \cite{chen2020generative}}                                                & 94.04 & 93.76 & 0.05 & 97.29 & 79.18  & 80.35 & 25.82 & 47.41 & 71.80 & 4.75  & 52.08  \\
\multicolumn{1}{c|}{}  & \multicolumn{1}{c|}{\textbf{LVM+POTGui (Ours)}}                                & \textbf{99.60} & \textbf{99.64} & \textbf{85.27}& \textbf{99.81} & \textbf{96.75}  & \textbf{99.05} & \textbf{82.26} & \textbf{93.85} & \textbf{96.43} & \textbf{85.97} & \textbf{91.64} \\ \hline
\end{tabularx}
\label{Tab:iGPT_tradi_classes}
\vspace{-0.5cm}
\end{table*}

\subsubsection{LVM+POTGui pk Benchmarks}
\label{iGPT_heads_vs_benchmarks}
In this experiment, our objective is to investigate the competitive edge of LVM+POTGui over current leading models. Specifically, we will benchmark the performance of LVM+POTGui against models such as BiSeNetV2 \cite{yu2021bisenet}, SegNet \cite{badrinarayanan2017segnet}, DeepLabv3+ \cite{chen2018encoderdecoder}, SegFormer \cite{xie2021segformer} and LVM \cite{chen2020generative}. Through this comparison, we aim to assess the convergence and performance of LVM+POTGui within the context of AD.

\Cref{Tab:iGPT_tradi} presents the average quantitative performance of LVM+POTGui in comparison to BiSeNetV2, SegNet, DeepLabv3+, SegFormer and LVM on both Cityscapes dataset and CamVid dataset. Analyzing this table, we can observe that LVM+POTGui consistently surpasses BiSeNetV2, SegNet, DeepLabv3+, SegFormer and LVM in all metrics on both datasets. For example, LVM+POTGui outperforms LVM by (99.99 - 45.81) / 45.81 = 118.27\% on Cityscapes dataset and (82.06 - 49.29) / 49.29 = 66.48\% on CamVid dataset. Such findings can be visually confirmed in \Cref{Fig:iGPT_tradi}. In addition, we can observe from \Cref{Fig:iGPT_tradi} that LVM+POTGui converges faster than all other competitors. For instance, on Cityscapes dataset (shown in \Cref{Fig:iGPT_tradi_e,Fig:iGPT_tradi_h}), LVM+POTGui converges at the first epoch while LVM converges at 16-th epoch in mIoU and mF1, resulting 15 times faster. Similarly, on CamVid dataset (illustrated in \Cref{Fig:iGPT_tradi_a,Fig:iGPT_tradi_d}), LVM+POTGui achieves around 6 times faster than LVM.

\Cref{Tab:iGPT_tradi_classes} delineates the class-wise performance of LVM+POTGui against other competitors on CamVid dataset. It reveals following distinct patterns: (I) Classes with extensive coverage or greater dimensions, such as Sky, Building, Road, Sidewalk, Tree, and Car, are well-inferred by nearly all models. These models demonstrate commendable accuracy, frequently surpassing 90\% across various metrics. (II) Conversely, for more slender classes like Pole and Fence, the performance of almost other competitors markedly decline to zero in all adopted metrics. Whereas LVM+POTGui maintains the best performance with over 80\% scores across all metrics. (III) In the case of classes with high shape variability, such as Bicyclist, the LVM+POTGui outperforms all competitors. This can be attributed to LVM's vast training dataset and POTGui's excellent optimization capability.

Moreover, \Cref{tab:semantic_pred} illustrates the prediction performance of the involved models for three real-world RGB images qualitatively. The findings still clearly demonstrates that LVM+POTGui outshines its competitors by achieving the best prediction in details. For example, on the one hand, the prediction of LVM+POTGui is segmented finer than others, such as the tree leafs; on the other hand, the slender pole is more approximate to the ground truth. 

\begin{table*}[tp]
\centering
\renewcommand{\arraystretch}{0.24}
\addtolength{\tabcolsep}{-0.45pt}
\caption{Inference performance comparison of semantic understanding driven by varieties of models on real-world dataset}
\begin{tabularx}{\linewidth}{|llllll|} 
\hline
\hspace{0.6cm}Raw Images &\hspace{0.3cm}Ground Truth &\hspace{0.3cm}BiSeNetV2 &\hspace{0.6cm}SegNet &\hspace{0.3cm}DeepLabv3+ &\hspace{-0.3cm}\textbf{LVM+POTGui (Ours)} \\
\hline
\includegraphics[width=0.162\linewidth, height=0.11\linewidth]{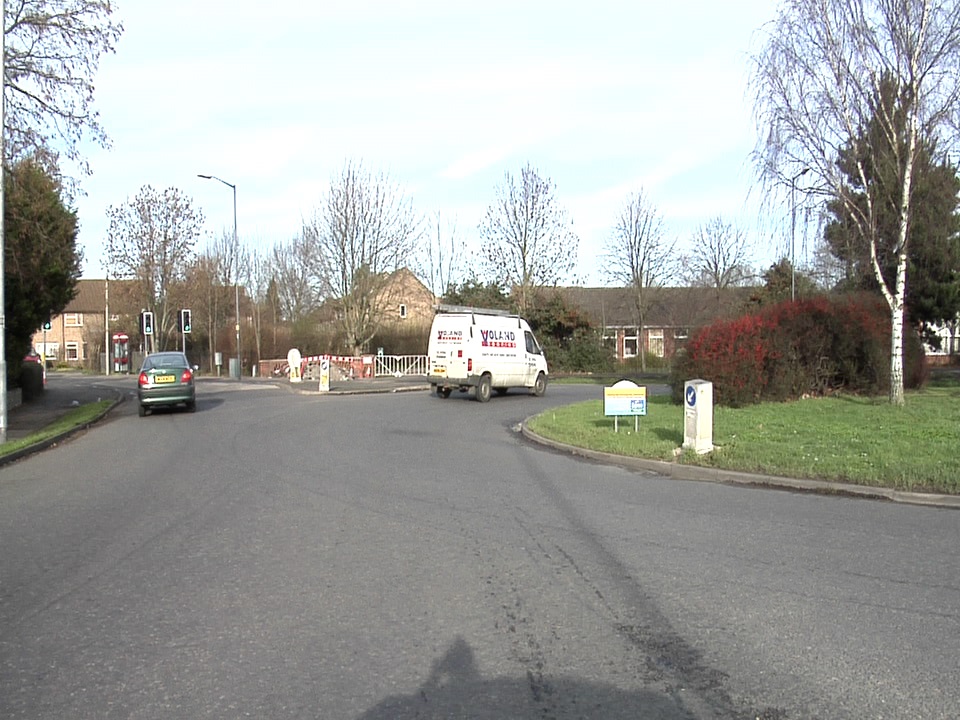} &\hspace{-0.47cm}
\includegraphics[width=0.162\linewidth, height=0.11\linewidth]{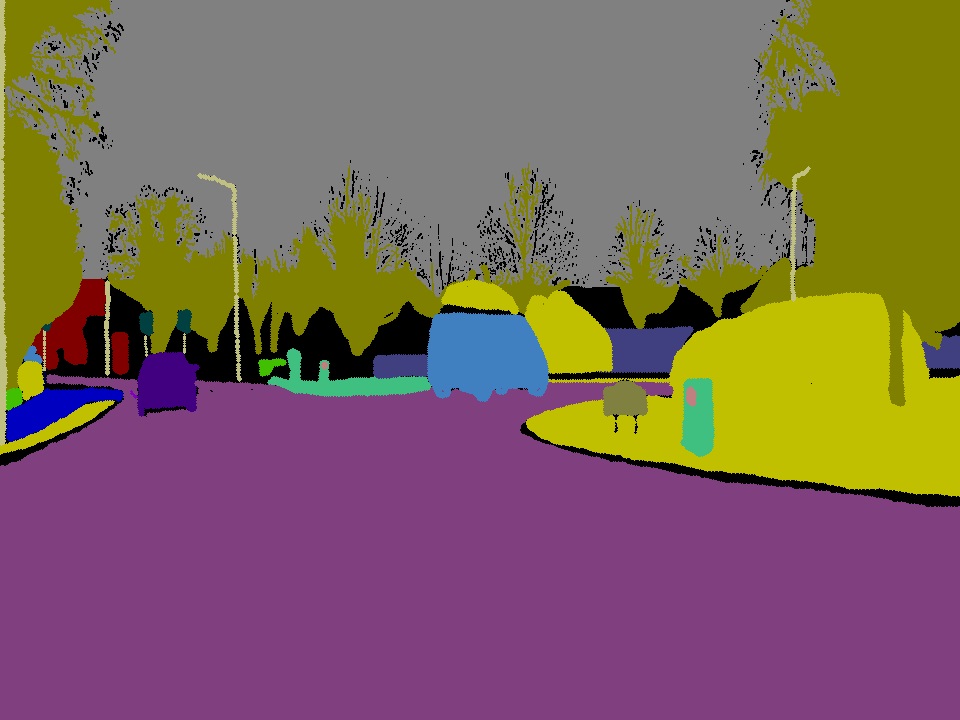} &\hspace{-0.47cm}
\includegraphics[width=0.162\linewidth, height=0.11\linewidth]{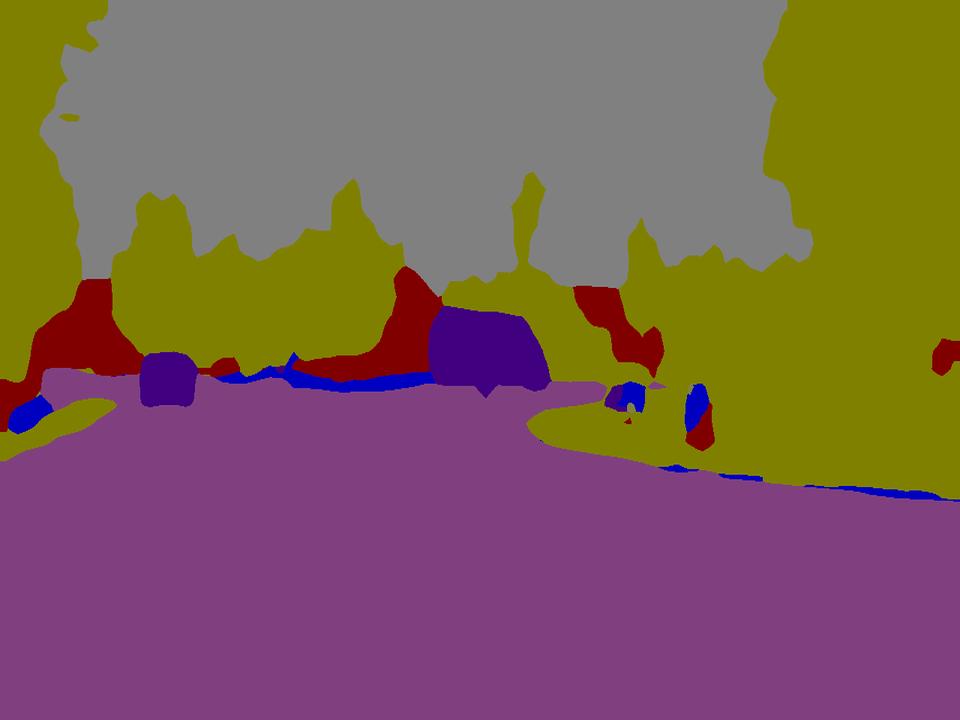} &\hspace{-0.47cm}
\includegraphics[width=0.162\linewidth, height=0.11\linewidth]{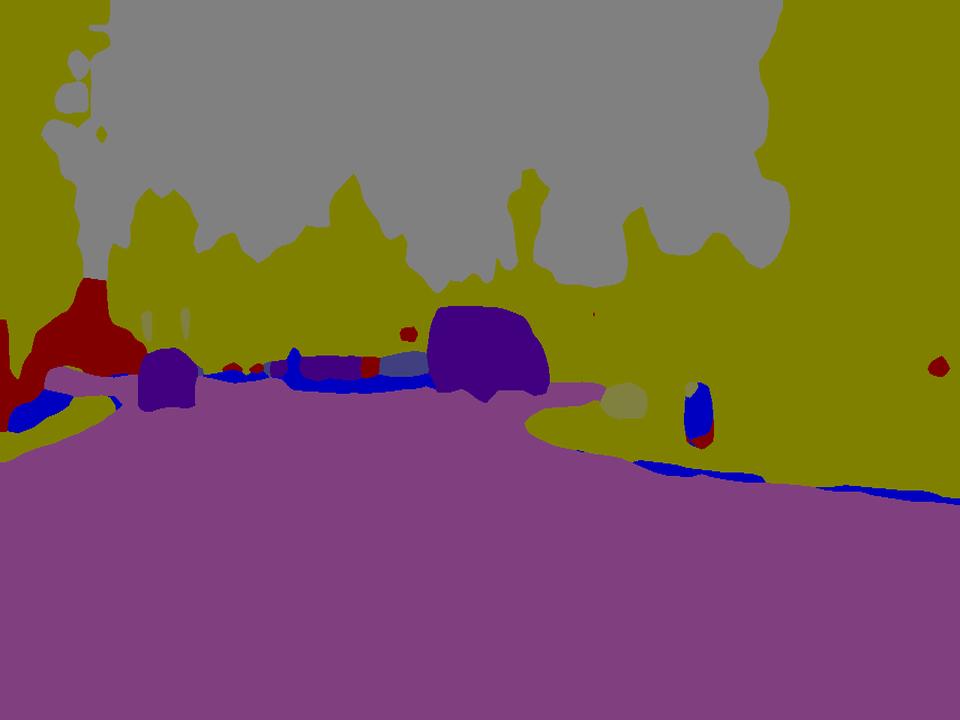} &\hspace{-0.47cm}
\includegraphics[width=0.162\linewidth, height=0.11\linewidth]{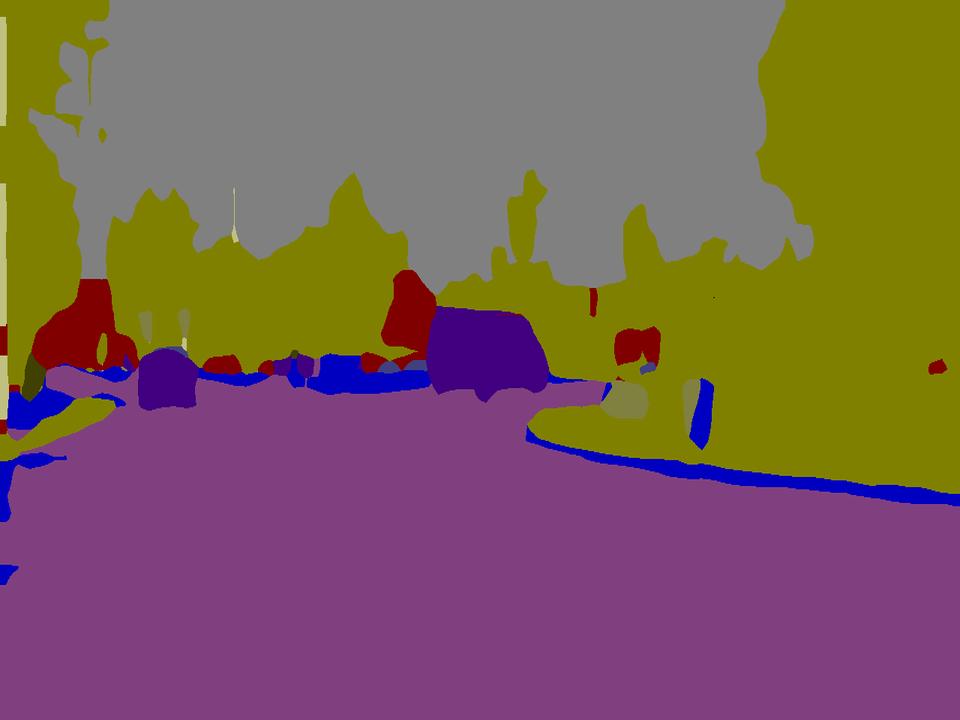} &\hspace{-0.47cm}
\includegraphics[width=0.162\linewidth, height=0.11\linewidth]{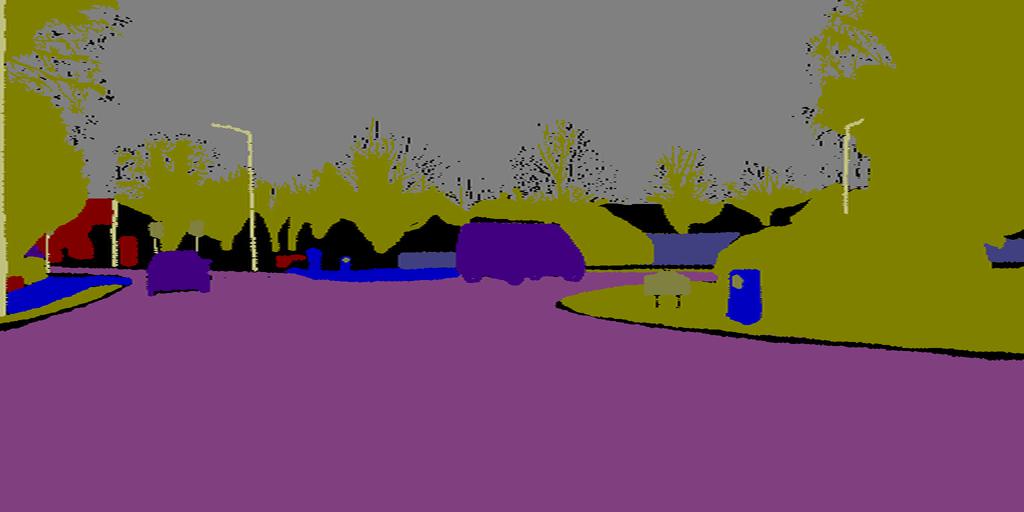}\\
\hline

\includegraphics[width=0.162\linewidth, height=0.11\linewidth]{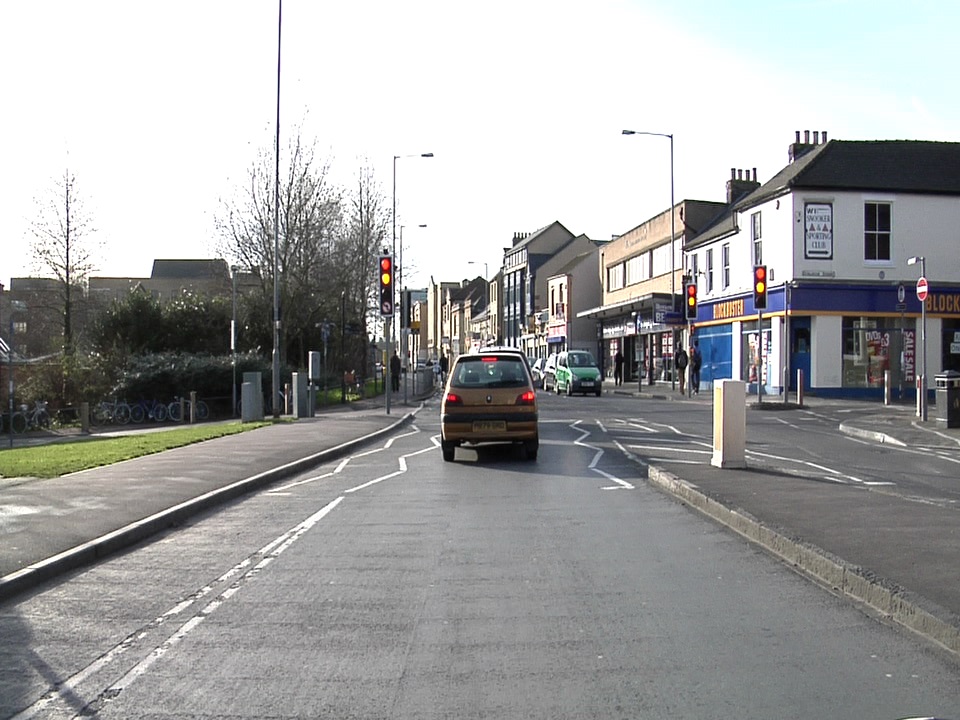} &\hspace{-0.47cm}
\includegraphics[width=0.162\linewidth, height=0.11\linewidth]{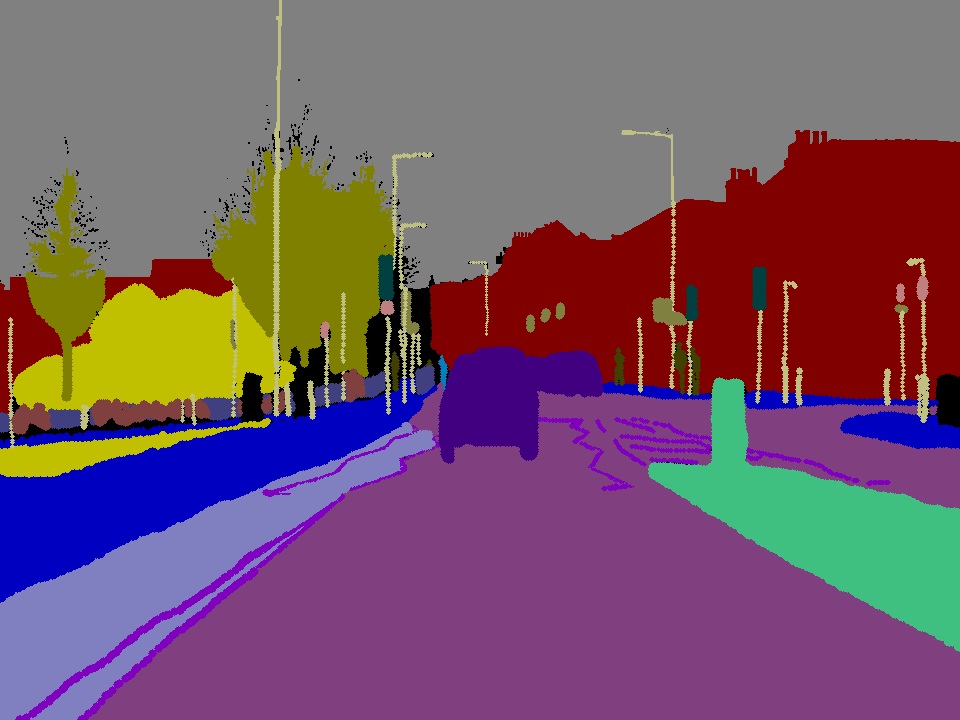} &\hspace{-0.47cm}
\includegraphics[width=0.162\linewidth, height=0.11\linewidth]{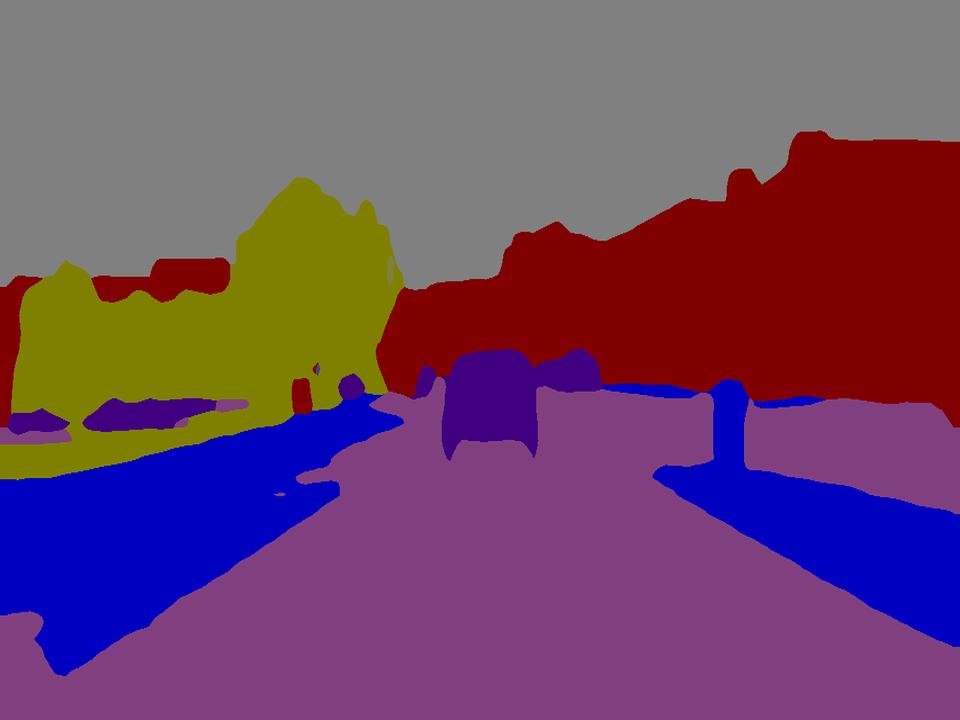} &\hspace{-0.47cm}
\includegraphics[width=0.162\linewidth, height=0.11\linewidth]{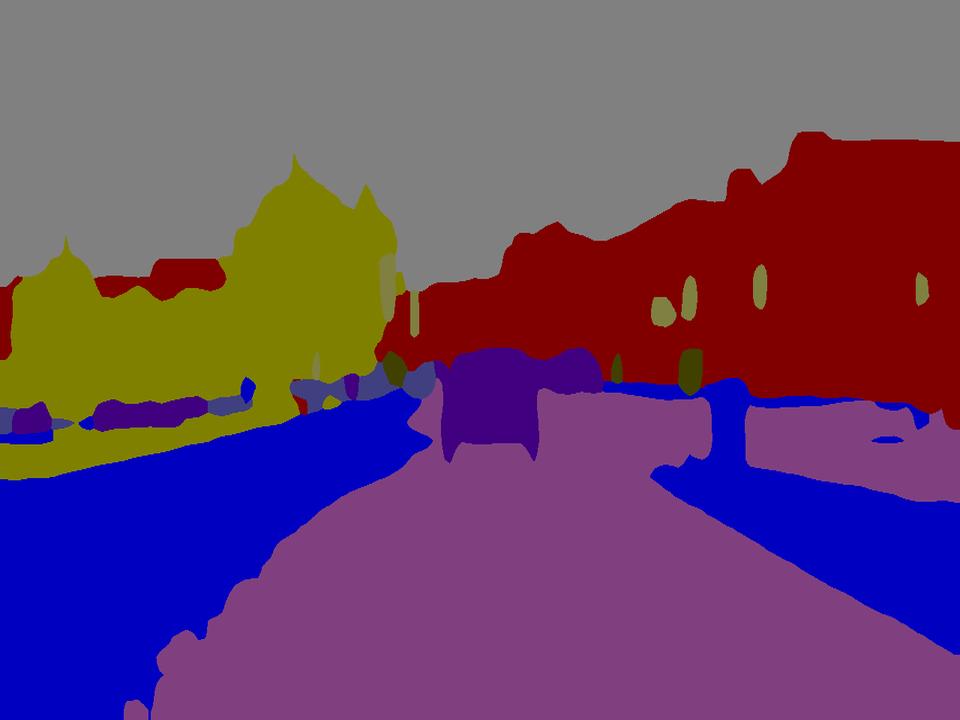} &\hspace{-0.47cm}
\includegraphics[width=0.162\linewidth, height=0.11\linewidth]{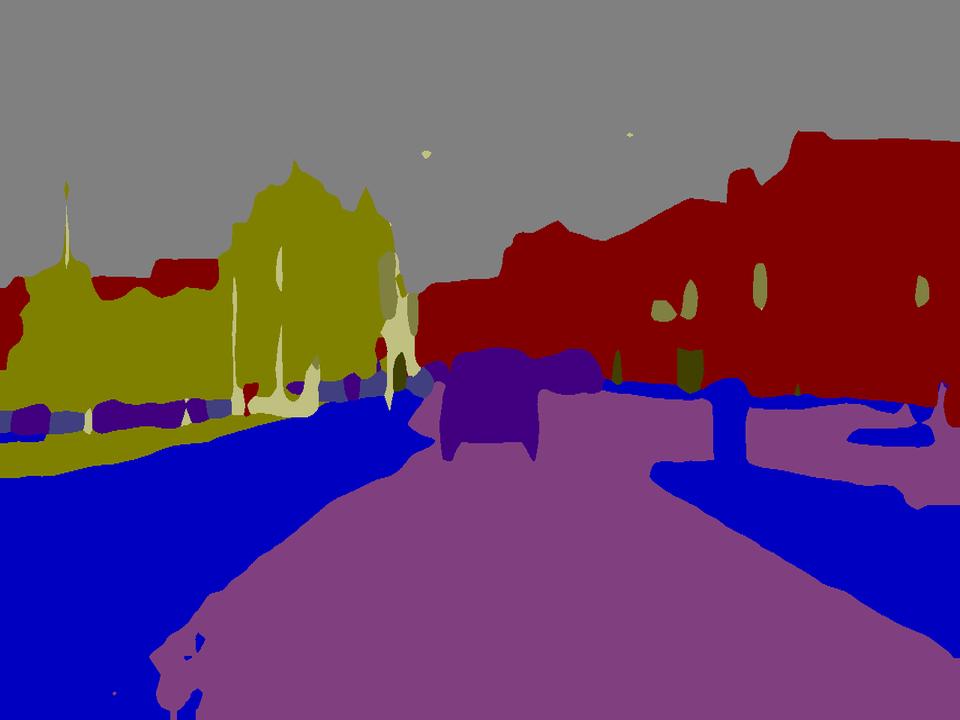} &\hspace{-0.47cm}
\includegraphics[width=0.162\linewidth, height=0.11\linewidth]{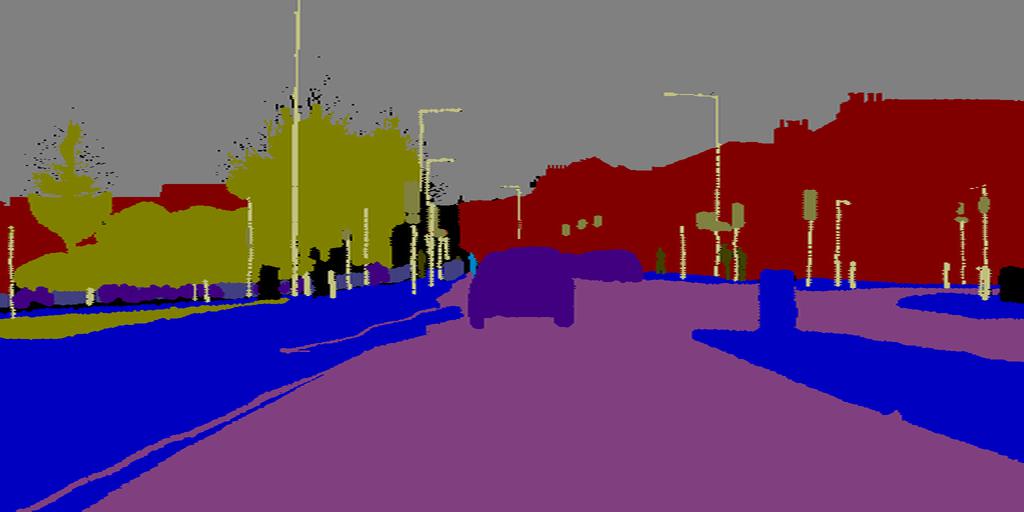}\\
\hline

\includegraphics[width=0.162\linewidth, height=0.11\linewidth]{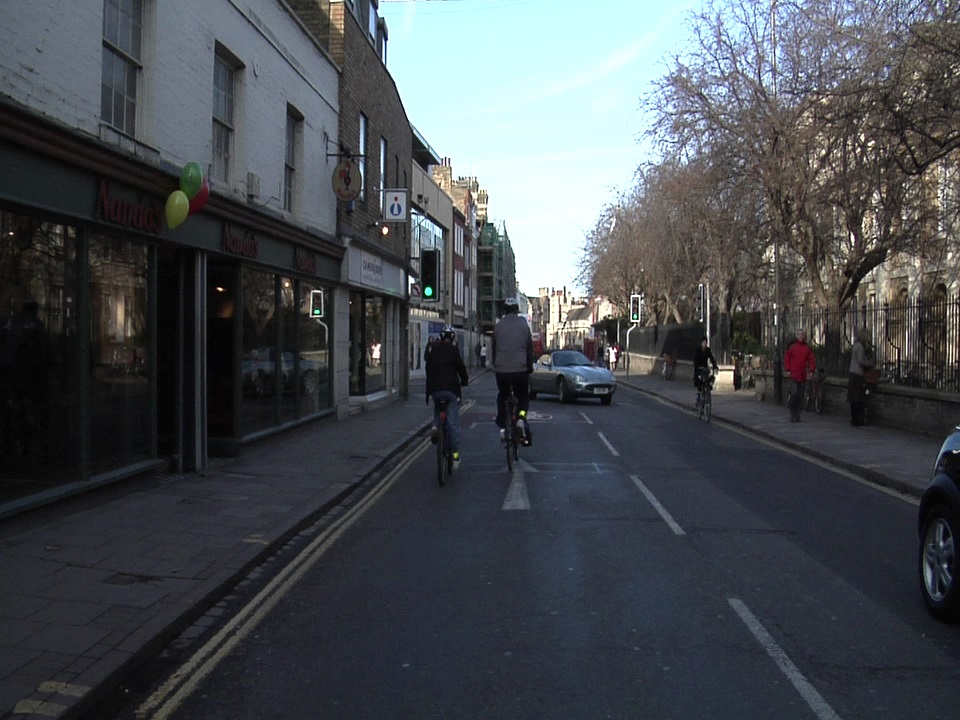} &\hspace{-0.47cm}
\includegraphics[width=0.162\linewidth, height=0.11\linewidth]{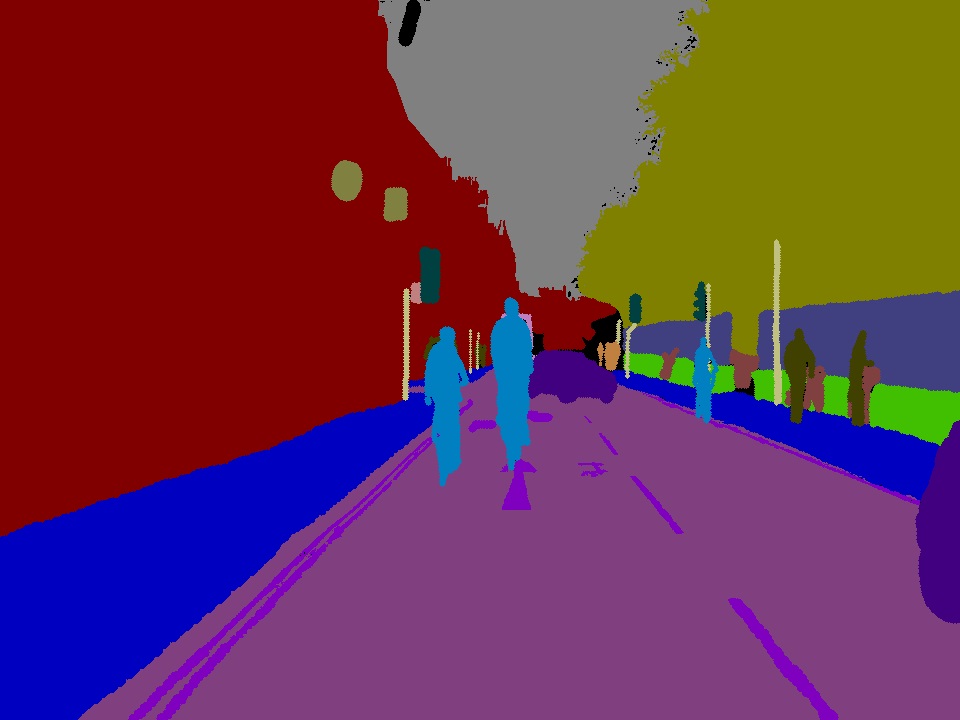} &\hspace{-0.47cm}
\includegraphics[width=0.162\linewidth, height=0.11\linewidth]{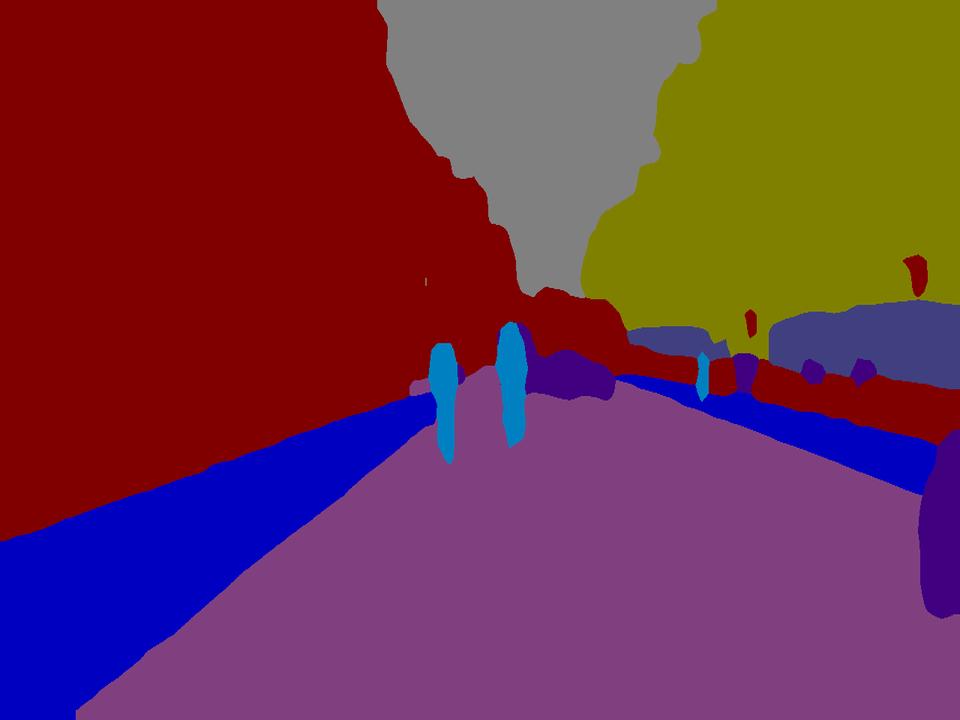} &\hspace{-0.47cm}
\includegraphics[width=0.162\linewidth, height=0.11\linewidth]{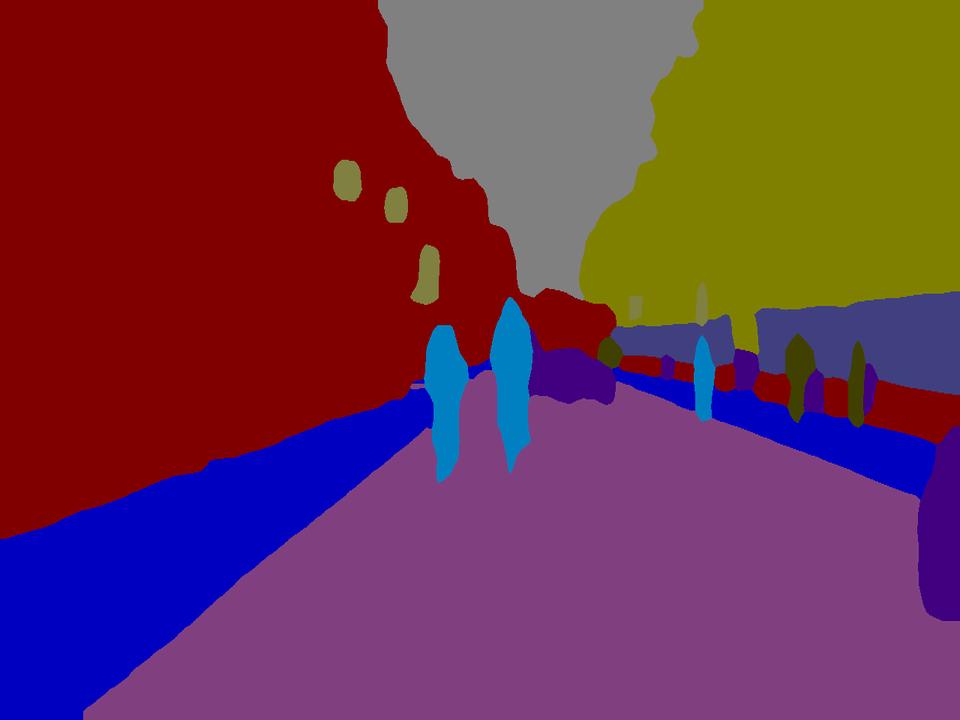} &\hspace{-0.47cm}
\includegraphics[width=0.162\linewidth, height=0.11\linewidth]{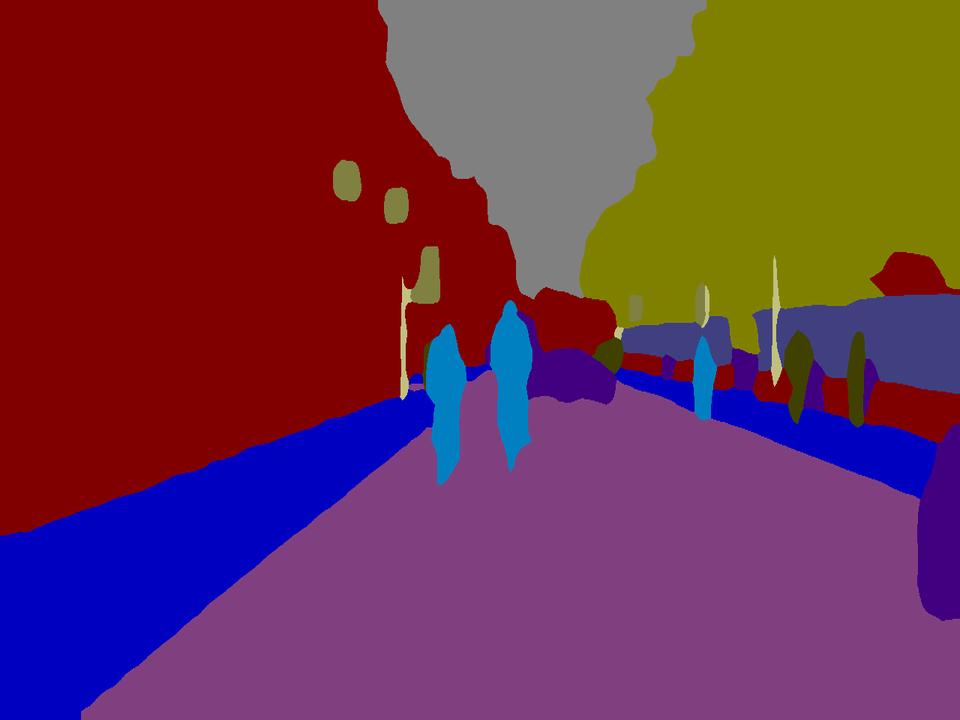} &\hspace{-0.47cm}
\includegraphics[width=0.162\linewidth, height=0.11\linewidth]{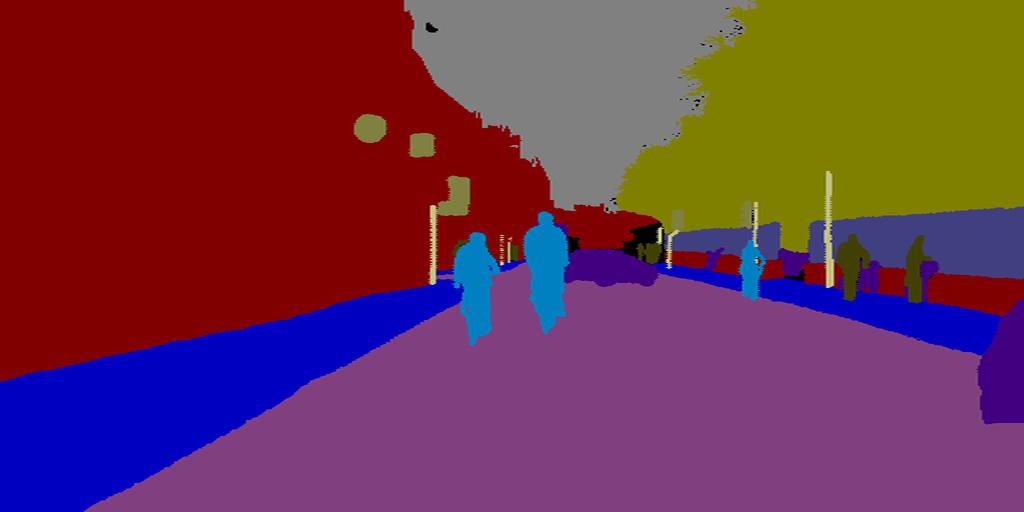}\\
\hline
\end{tabularx}
\label{tab:semantic_pred}
\vspace{-0.4cm}
\end{table*}

\subsection{Ablation Study}
\begin{figure}[tp]
\vspace{-0.3cm}
\centering
\subfloat[mIoU]{\includegraphics[width=0.5\linewidth, height=0.4\linewidth]{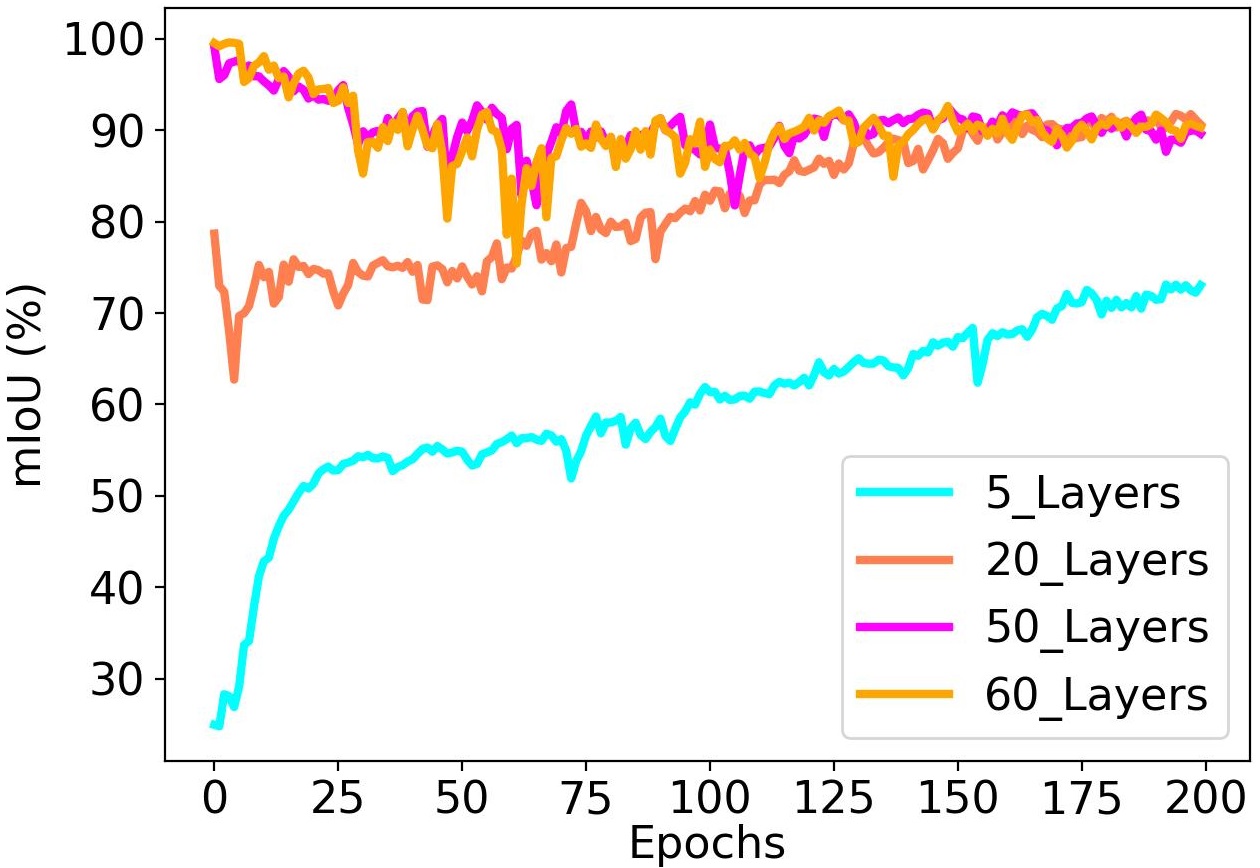}%
\label{Fig:layers_mIoU}}
\subfloat[mF1]{\includegraphics[width=0.5\linewidth, height=0.4\linewidth]{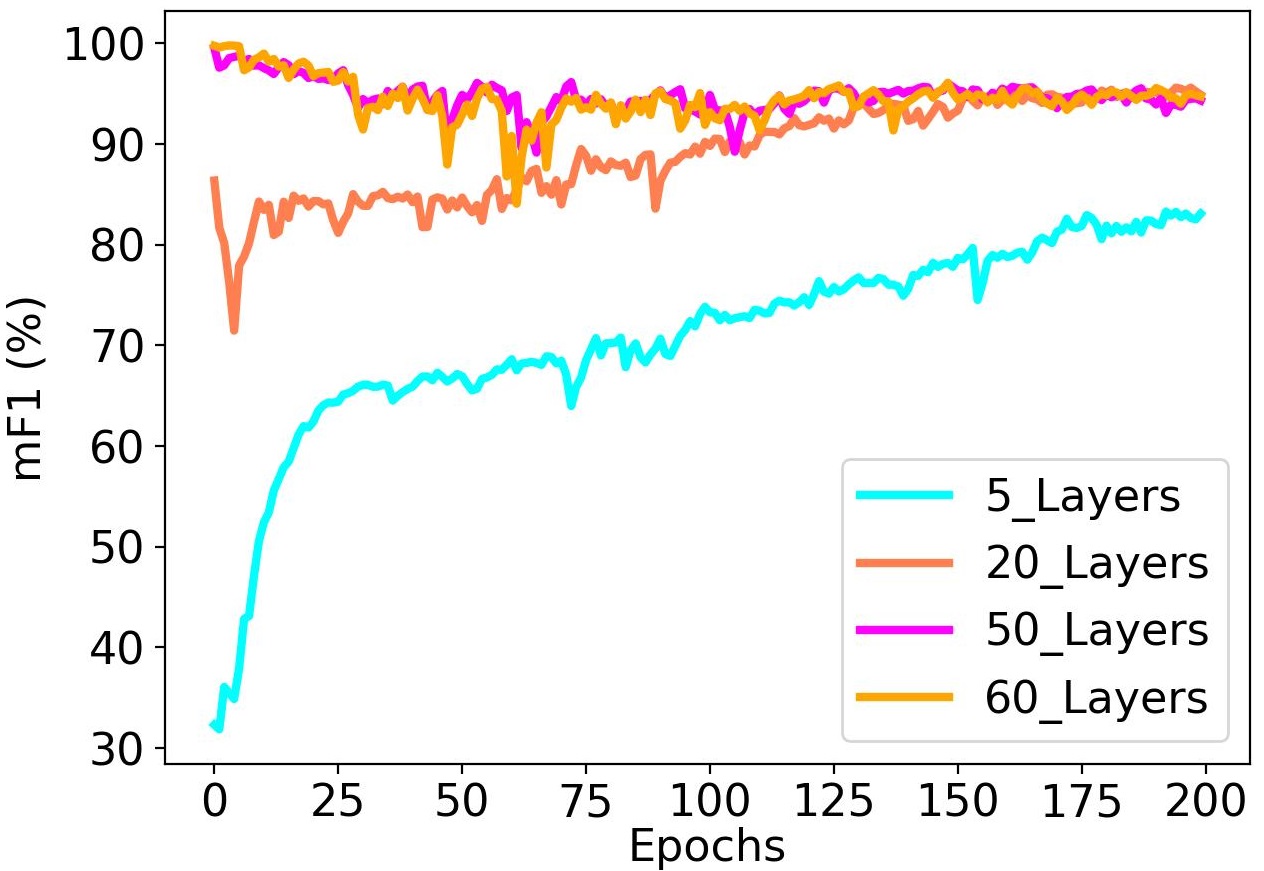}%
\label{Fig:layers_mF1}}
\caption{The effect of the number of layers on POTGui.}
\label{Fig:POTGen_layers}
\end{figure}

The number of layers in POTGen affects the performance of the proposed POTGui optimization scheme. \Cref{Fig:POTGen_layers} compares the performance of different cases of POTGui with different number of layers. We can observe following patterns from \Cref{Fig:POTGen_layers}: (I) The more layers POTGen contains, the better performance LVM+POTGui has. (II) The more layers the POTGen has, the faster LVM+POTGui converges. (III) When the number of layers of POTGen exceeds a certain value, the performance of LVM+POTGui does not improve any more. For example, the case of 50 layers performs almost same with the case of 60 layers.

\subsection{Real Driving Test}
\begin{figure}[tp]
\centering
\subfloat[Real Driving Test on Apolloscapes]{\includegraphics[width=\linewidth,height=0.47\linewidth]{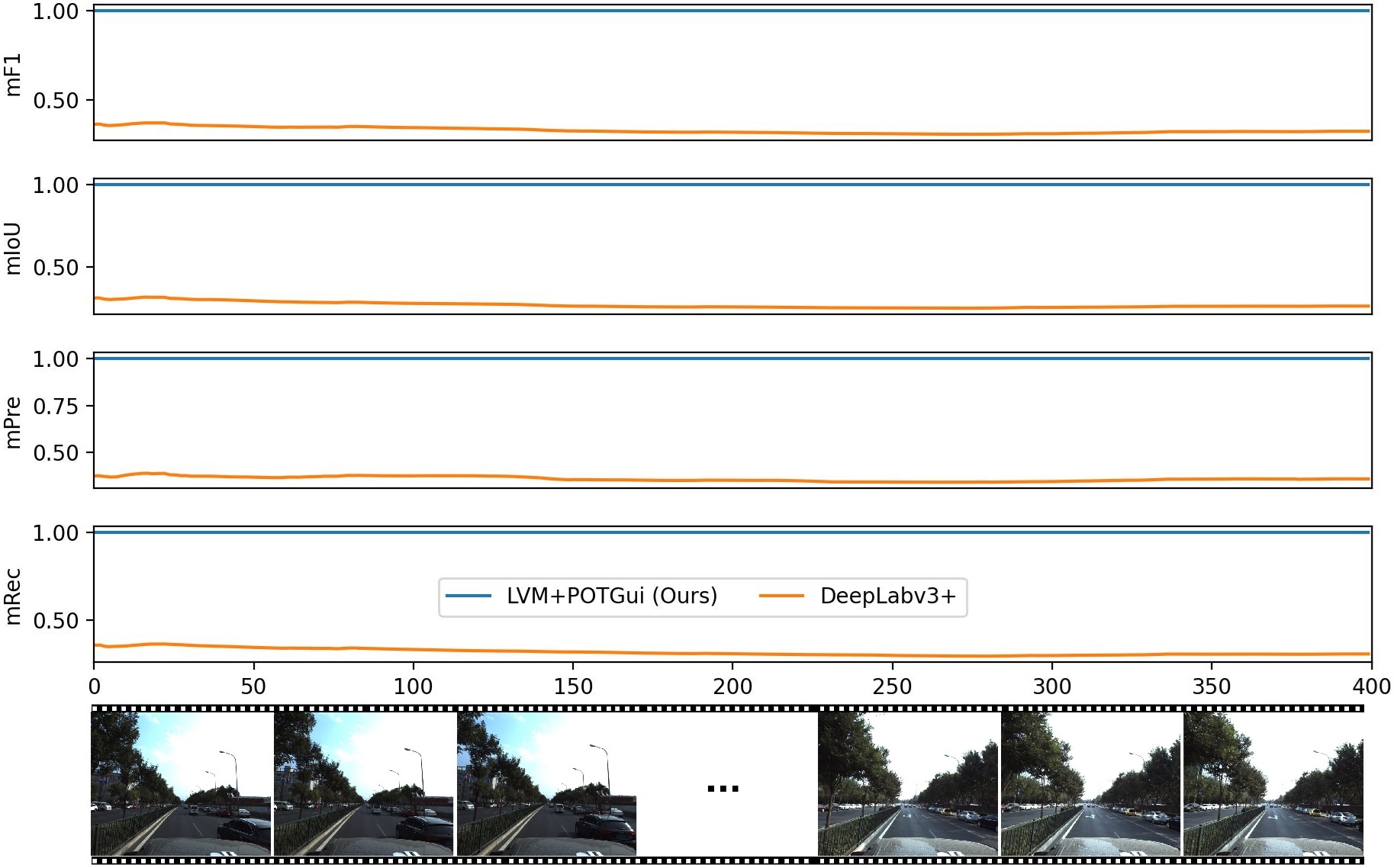}%
\label{Fig:apollo_inf}}

\subfloat[Real Driving Test on CARLA\_ADV]{\includegraphics[width=\linewidth,height=0.47\linewidth]{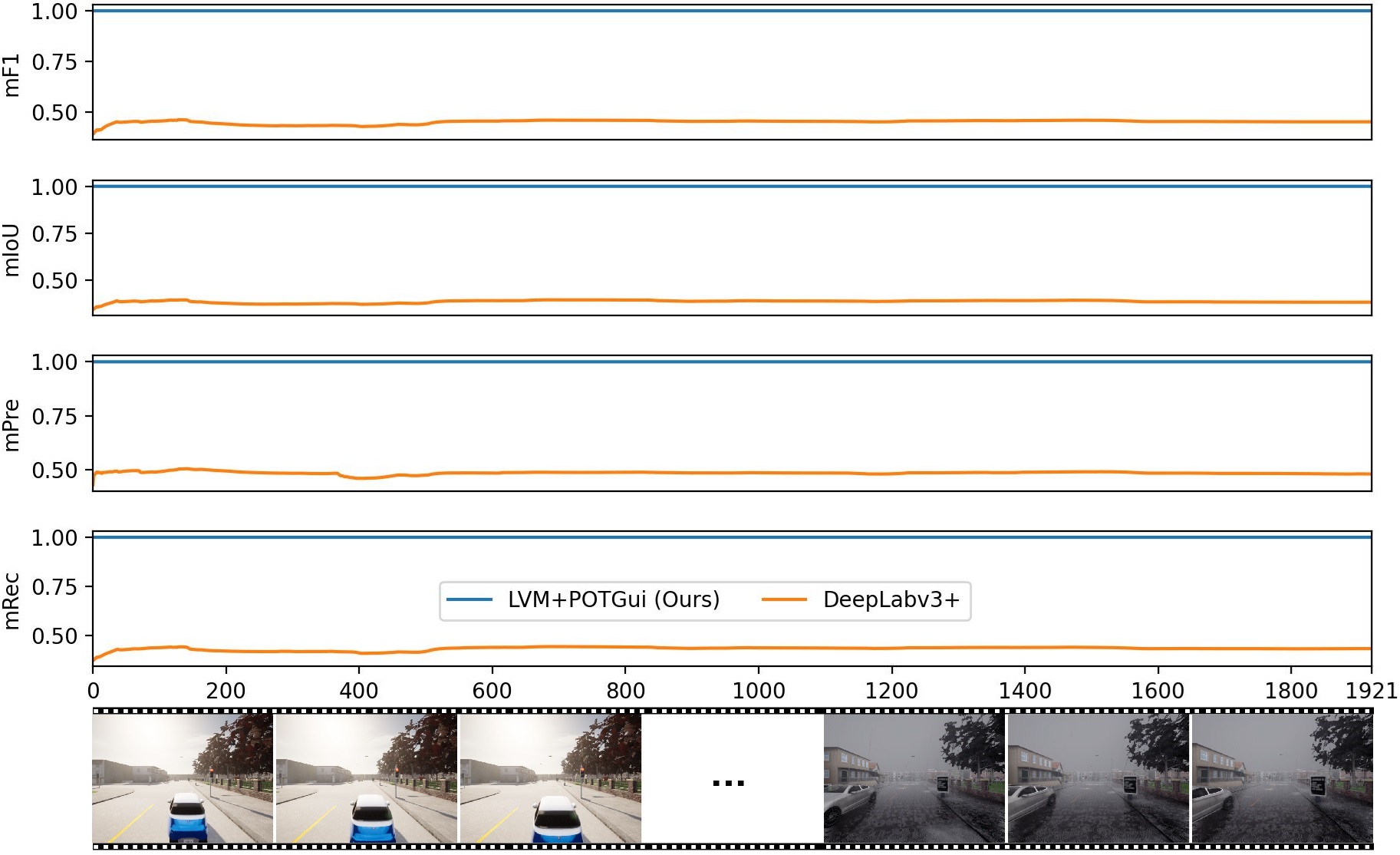}%
\label{Fig:carla_adv_inf}}
\caption{Real driving test of LVM+POTGui on Apolloscapes dataset and CARLA\_ADV dataset.}
\label{Fig:driving_inf}
\vspace{-0.1cm}
\end{figure}

\Cref{Fig:driving_inf} compares the inference performance of LVM+POTGui against DeepLabv3+ (best performing competitor) in real driving test. Notably, LVM+POTGui consistently demonstrates better performance relative to DeepLabv3+, which can be supported by following aspects: (I) LVM+POTGui consistently achieves higher scores for all metrics across all sequential frames, indicating better accuracy. (II) LVM+POTGui shows small performance variance across varied test conditions, such as fog, cloudy, rainy, dark, and  combinations thereof, suggesting robustness against adverse weather conditions.

\section{Conclusion}
This paper introduced a POTGui-endowed LVM-driven street scene semantic understanding method in the context of AD. It involved deploying LVM and POTGui optimization scheme on vehicle to understand semantic information of driving surroundings. We carried out comprehensive experiments on real-world dataset (\ie, Cityscapes dataset, CamVid dataset and ApolloScapes dataset) and deployed the proposed method in CARLA simulation platform to verify it. Experimental results demonstrated that the proposed method outperforms existing approaches. Future work plans to incorporate multi-modal data into the proposed method, such as Lidar and depth camera.


\begin{thebibliography}{10}
\bibitem{10416354}
J.~Rückin, F.~Magistri, C.~Stachniss, and M.~Popović, ``Semi-supervised active learning for semantic segmentation in unknown environments using informative path planning,'' \emph{IEEE Robotics and Automation Letters}, vol.~9, no.~3, pp. 2662--2669, 2024.

\bibitem{10342110}
H.~Son and J.~Weiland, ``Lightweight semantic segmentation network for semantic scene understanding on low-compute devices,'' in \emph{2023 IEEE/RSJ International Conference on Intelligent Robots and Systems (IROS)}, 2023, pp. 62--69.

\bibitem{10342254}
Z.~Chen, Z.~Ding, J.~M. Gregory, and L.~Liu, ``Ida: Informed domain adaptive semantic segmentation,'' in \emph{2023 IEEE/RSJ International Conference on Intelligent Robots and Systems(IROS)}, 2023, pp. 90--97.

\bibitem{10341639}
J.~Li, W.~Shi, D.~Zhu, G.~Zhang, X.~Zhang, and J.~Li, ``Featdanet: Feature-level domain adaptation network for semantic segmentation,'' in \emph{2023 IEEE/RSJ International Conference on Intelligent Robots and Systems (IROS)}, 2023, pp. 3873--3880.

\bibitem{kou2024fedrc}
W.-B. Kou, Q.~Lin, M.~Tang, S.~Wang, G.~Zhu, and Y.-C. Wu, ``Fedrc: A rapid-converged hierarchical federated learning framework in street scene semantic understanding,'' in \emph{2024 IEEE/RSJ International Conference on Intelligent Robots and Systems (IROS)}.\hskip 1em plus 0.5em minus 0.4em\relax IEEE, 2024, pp. 2578--2585.

\bibitem{kou2024fast}
W.-B. Kou, Q.~Lin, M.~Tang, R.~Ye, S.~Wang, G.~Zhu, and Y.-C. Wu, ``Fast-convergent and communication-alleviated heterogeneous hierarchical federated learning in autonomous driving,'' \emph{arXiv preprint arXiv:2409.19560}, 2024.

\bibitem{kou2024adverse}
W.-B. Kou, G.~Zhu, R.~Ye, S.~Wang, Q.~Lin, M.~Tang, and Y.-C. Wu, ``An adverse weather-immune scheme with unfolded regularization and foundation model knowledge distillation for street scene understanding,'' \emph{arXiv preprint arXiv:2409.14737}, 2024.

\bibitem{10160999}
N.~Kim, T.~Son, J.~Pahk, C.~Lan, W.~Zeng, and S.~Kwak, ``Wedge: Web-image assisted domain generalization for semantic segmentation,'' in \emph{2023 IEEE International Conference on Robotics and Automation (ICRA)}, 2023, pp. 9281--9288.

\bibitem{9811702}
J.~Tian, N.~C. Mithun, Z.~Seymour, H.-P. Chiu, and Z.~Kira, ``Striking the right balance: Recall loss for semantic segmentation,'' in \emph{2022 International Conference on Robotics and Automation (ICRA)}, 2022, pp. 5063--5069.

\bibitem{10610948}
J.~Sun, Q.~Zhang, Y.~Duan, X.~Jiang, C.~Cheng, and R.~Xu, ``Prompt, plan, perform: Llm-based humanoid control via quantized imitation learning,'' in \emph{2024 IEEE International Conference on Robotics and Automation (ICRA)}, 2024, pp. 16\,236--16\,242.

\bibitem{10611614}
Y.~Duan, Q.~Zhang, and R.~Xu, ``Prompting multi-modal tokens to enhance end-to-end autonomous driving imitation learning with llms,'' in \emph{2024 IEEE International Conference on Robotics and Automation (ICRA)}, 2024, pp. 6798--6805.

\bibitem{Zhu2023mini}
D.~Zhu, J.~Chen, X.~Shen, X.~Li, and M.~Elhoseiny, ``Minigpt-4: Enhancing vision-language understanding with advanced large language models,'' \emph{CoRR}, vol. abs/2304.10592, 2023.

\bibitem{ouyang2022training}
L.~Ouyang, J.~Wu, X.~Jiang, D.~Almeida, C.~L. Wainwright, P.~Mishkin, C.~Zhang, S.~Agarwal, K.~Slama, A.~Ray, J.~Schulman, J.~Hilton, F.~Kelton, L.~Miller, M.~Simens, A.~Askell, P.~Welinder, P.~F. Christiano, J.~Leike, and R.~Lowe, ``Training language models to follow instructions with human feedback,'' in \emph{NeurIPS}, 2022.

\bibitem{jiang2023structgpt}
J.~Jiang, K.~Zhou, Z.~Dong, K.~Ye, W.~X. Zhao, and J.-R. Wen, ``Structgpt: A general framework for large language model to reason over structured data,'' 2023.

\bibitem{wei2022chain}
J.~Wei, X.~Wang, D.~Schuurmans, M.~Bosma, B.~Ichter, F.~Xia, E.~H. Chi, Q.~V. Le, and D.~Zhou, ``Chain-of-thought prompting elicits reasoning in large language models,'' in \emph{NeurIPS}, 2022.

\bibitem{huang2023visual}
C.~Huang, O.~Mees, A.~Zeng, and W.~Burgard, ``Visual language maps for robot navigation,'' in \emph{2023 IEEE International Conference on Robotics and Automation (ICRA)}.\hskip 1em plus 0.5em minus 0.4em\relax IEEE, 2023, pp. 10\,608--10\,615.

\bibitem{10388394}
Y.~Yang, Z.~Zhou, J.~Wu, Y.~Wang, and R.~Xiong, ``Class semantics modulation for open-set instance segmentation,'' \emph{IEEE Robotics and Automation Letters}, vol.~9, no.~3, pp. 2240--2247, 2024.

\bibitem{9968085}
X.~Liu, Y.~Zhang, and D.~Shan, ``Unseen object few-shot semantic segmentation for robotic grasping,'' \emph{IEEE Robotics and Automation Letters}, vol.~8, no.~1, pp. 320--327, 2023.

\bibitem{10049523}
Z.~Feng, Y.~Guo, and Y.~Sun, ``Cekd: Cross-modal edge-privileged knowledge distillation for semantic scene understanding using only thermal images,'' \emph{IEEE Robotics and Automation Letters}, vol.~8, no.~4, pp. 2205--2212, 2023.

\bibitem{10168231}
A.~Almin, L.~Lemarié, A.~Duong, and B.~R. Kiran, ``Navya3dseg - navya 3d semantic segmentation dataset design \& split generation for autonomous vehicles,'' \emph{IEEE Robotics and Automation Letters}, vol.~8, no.~9, pp. 5584--5591, 2023.

\bibitem{10161421}
Q.~Yan, S.~Li, C.~Liu, M.~Liu, and Q.~Chen, ``Fdlnet: Boosting real-time semantic segmentation by image-size convolution via frequency domain learning,'' in \emph{2023 IEEE International Conference on Robotics and Automation (ICRA)}, 2023.

\bibitem{yang2022deaot}
Z.~Yang and Y.~Yang, ``Decoupling features in hierarchical propagation for video object segmentation,'' in \emph{Advances in Neural Information Processing Systems (NeurIPS)}, 2022.

\bibitem{zhou2022rethinking}
T.~Zhou, W.~Wang, E.~Konukoglu, and L.~Van~Gool, ``Rethinking semantic segmentation: A prototype view,'' in \emph{Proceedings of the IEEE/CVF Conference on Computer Vision and Pattern Recognition}, 2022, pp. 2582--2593.

\bibitem{xie2021segformer}
E.~Xie, W.~Wang, Z.~Yu, A.~Anandkumar, J.~M. Alvarez, and P.~Luo, ``Segformer: Simple and efficient design for semantic segmentation with transformers,'' in \emph{Neural Information Processing Systems (NeurIPS)}, 2021.

\bibitem{9697426}
Y.~B. Can, A.~Liniger, O.~Unal, D.~Paudel, and L.~Van~Gool, ``Understanding bird’s-eye view of road semantics using an onboard camera,'' \emph{IEEE Robotics and Automation Letters}, vol.~7, no.~2, pp. 3302--3309, 2022.

\bibitem{feddrive2022}
L.~Fantauzzo, E.~Fanì, D.~Caldarola, A.~Tavera, F.~Cermelli, M.~Ciccone, and B.~Caputo, ``Feddrive: Generalizing federated learning to semantic segmentation in autonomous driving,'' in \emph{Proceedings of the 2022 IEEE/RSJ International Conference on Intelligent Robots and Systems}, 2022.

\bibitem{10342134}
W.-B. Kou, S.~Wang, G.~Zhu, B.~Luo, Y.~Chen, D.~W. Kwan~Ng, and Y.-C. Wu, ``Communication resources constrained hierarchical federated learning for end-to-end autonomous driving,'' in \emph{2023 IEEE/RSJ International Conference on Intelligent Robots and Systems (IROS)}, 2023, pp. 9383--9390.

\bibitem{chen2020generative}
M.~Chen, A.~Radford, R.~Child, J.~Wu, H.~Jun, D.~Luan, and I.~Sutskever, ``Generative pretraining from pixels,'' in \emph{International conference on machine learning}.\hskip 1em plus 0.5em minus 0.4em\relax PMLR, 2020, pp. 1691--1703.

\bibitem{chen2017deeplab}
L.-C. Chen, G.~Papandreou, I.~Kokkinos, K.~Murphy, and A.~L. Yuille, ``Deeplab: Semantic image segmentation with deep convolutional nets, atrous convolution, and fully connected crfs,'' \emph{IEEE transactions on pattern analysis and machine intelligence}, vol.~40, no.~4, pp. 834--848, 2017.

\bibitem{10529194}
Q.~Lin, Y.~Li, W.-B. Kou, T.-H. Chang, and Y.-C. Wu, ``Communication-efficient activity detection for cell-free massive mimo: An augmented model-driven end-to-end learning framework,'' \emph{IEEE Transactions on Wireless Communications}, pp. 1--1, 2024.

\bibitem{lin2023communication}
Q.~Lin, Y.~Li, W.-B. Kou, T.~Chang, and Y.-C. Wu, ``Communication-efficient joint signal compression and activity detection in cell-free massive mimo,'' in \emph{ICC 2023-IEEE International Conference on Communications}.\hskip 1em plus 0.5em minus 0.4em\relax IEEE, 2023, pp. 5030--5035.

\bibitem{Cordts2016Cityscapes}
M.~Cordts, M.~Omran, S.~Ramos, T.~Rehfeld, M.~Enzweiler, R.~Benenson, U.~Franke, S.~Roth, and B.~Schiele, ``The cityscapes dataset for semantic urban scene understanding,'' in \emph{Proc. of the IEEE Conference on Computer Vision and Pattern Recognition (CVPR)}, 2016.

\bibitem{brostow2008segmentation}
G.~J. Brostow, J.~Shotton, J.~Fauqueur, and R.~Cipolla, ``Segmentation and recognition using structure from motion point clouds,'' in \emph{Computer Vision--ECCV 2008: 10th European Conference on Computer Vision, Marseille, France, October 12-18, 2008, Proceedings, Part I 10}.\hskip 1em plus 0.5em minus 0.4em\relax Springer, 2008, pp. 44--57.

\bibitem{wang2019apolloscape}
P.~Wang, X.~Huang, X.~Cheng, D.~Zhou, Q.~Geng, and R.~Yang, ``The apolloscape open dataset for autonomous driving and its application,'' \emph{IEEE transactions on pattern analysis and machine intelligence}, 2019.

\bibitem{dosovitskiy2017carla}
A.~Dosovitskiy, G.~Ros, F.~Codevilla, A.~Lopez, and V.~Koltun, ``Carla: An open urban driving simulator,'' in \emph{Proceedings of The 1st Annual Conference on Robot Learning}, Oct. 2017, pp. 1--16.

\bibitem{yu2021bisenet}
C.~Yu, C.~Gao, J.~Wang, G.~Yu, C.~Shen, and N.~Sang, ``Bisenet v2: Bilateral network with guided aggregation for real-time semantic segmentation,'' \emph{International Journal of Computer Vision}, vol. 129, pp. 3051--3068, 2021.

\bibitem{badrinarayanan2017segnet}
V.~Badrinarayanan, A.~Kendall, and R.~Cipolla, ``Segnet: A deep convolutional encoder-decoder architecture for image segmentation,'' \emph{IEEE transactions on pattern analysis and machine intelligence}, vol.~39, no.~12, pp. 2481--2495, 2017.

\bibitem{chen2018encoderdecoder}
L.-C. Chen, Y.~Zhu, G.~Papandreou, F.~Schroff, and H.~Adam, ``Encoder-decoder with atrous separable convolution for semantic image segmentation,'' 2018.

\end{thebibliography}
\end{document}